\documentclass{article}

     \PassOptionsToPackage{numbers}{natbib}

     \usepackage[final]{neurips_2021}

\usepackage[utf8]{inputenc} 
\usepackage[T1]{fontenc}    
\usepackage{hyperref}       
\usepackage{url}            
\usepackage{booktabs}       
\usepackage{amsfonts}       
\usepackage{nicefrac}       
\usepackage{microtype}      
\usepackage{xcolor}         

\usepackage[numbers]{natbib}
\usepackage{graphicx}
\usepackage{caption}
\usepackage{amsmath}
\usepackage{hyperref}
\usepackage{wrapfig}
\usepackage{booktabs}
\usepackage{multirow}
\usepackage{lscape}
\usepackage{algorithmic}
\usepackage[linesnumbered,ruled,vlined]{algorithm2e}
\usepackage{threeparttable}
\usepackage{enumitem}

\usepackage{subfigure}
\usepackage{longtable}
\usepackage{array}

\setcitestyle{comma}

\title{Learning Large Neighborhood Search Policy for Integer Programming}

\author{%
  Yaoxin~Wu\\
  SCALE@NTU  Corp  Lab\\
  Nanyang  Technological University, Singapore \\
  \texttt{wuyaoxin@ntu.edu.sg} \\
  \And
  Wen Song\thanks{Wen Song is the corresponding author.} \\
  Shandong University\\
  Qingdao, China \\
  \texttt{wensong@email.sdu.edu.cn} \\
  \AND
  Zhiguang Cao \\
  Singapore Institute of Manufacturing Technology\\
  A*STAR, Singapore \\
  \texttt{zhiguangcao@outlook.com} \\
  \And
  Jie Zhang \\
  Nanyang Technological University \\
  Singapore \\
  \texttt{zhangj@ntu.edu.sg} \\
}

\begin{document}

\maketitle

\begin{abstract}
We propose a deep reinforcement learning (RL) method to learn large neighborhood search (LNS) policy for integer programming (IP). The RL policy is trained as the destroy operator to select a subset of variables at each step, which is reoptimized by an IP solver as the repair operator. However, the combinatorial number of variable subsets prevents direct application of typical RL algorithms. To tackle this challenge, we represent all subsets by factorizing them into binary decisions on each variable. We then design a neural network to learn policies for each variable in parallel, trained by a customized actor-critic algorithm. We evaluate the proposed method on four representative IP problems. Results show that it can find better solutions than SCIP in much less time, and significantly outperform other LNS baselines with the same runtime. Moreover, these advantages notably persist when the policies generalize to larger problems. Further experiments with Gurobi also reveal that our method can outperform this state-of-the-art commercial solver within the same time limit.
\end{abstract}

\section{Introduction}

Combinatorial optimization problems (COPs) have been widely studied in computer science and operations research, which cover numerous real-world tasks in many fields such as communication, transportation and manufacturing \cite{paschos2013applications}. Most COPs are very difficult to solve efficiently due to their NP-hardness. The performance of classic methods, including exact and heuristic algorithms \cite{papadimitriou1998combinatorial}, is generally limited by hand-crafted policies that are costly to design, since considerable trial-and-error and domain knowledge are needed. On the other hand, it is common in practice that similar instances with shared structure are frequently solved, and differ only in data that normally follows a distribution \cite{bengio2018machine}. This provides a chance for machine learning to \emph{automatically} generate heuristics or policies. In doing so, the learned alternatives are expected to save massive manual work in algorithm design, and raise the performance of the algorithm on a class of problems. 

Recently, a number of works apply deep (reinforcement) learning to automatically design heuristic algorithms, either in constructive or improving fashion. Different from construction heuristics that sequentially extend partial solutions to complete ones \cite{vinyals2015pointer,Bello2017WorkshopT,khalil2017learning,kool2018attention,xin2020multi,li2021deep,li2021heterogeneous,zhang2020learning}, learning \emph{improvement heuristics} can often deliver high solution quality by iteratively reoptimizing an initial solution using local operations \cite{chen2019learning,wu2019learning,lu2019learning}. In this line, some methods are developed under the Large Neighborhood Search (LNS) framework \cite{hottung2019neural,gao2020learn}, which is a powerful improving paradigm to find near-optimal solutions for COPs. 


However, the above methods are restricted to specific problem types, and cannot generalize to those from different domains. This motivates the studies of learning to directly solve Integer Programs (IPs), which is very powerful and flexible in modelling a wide range of COPs. The standard approach to solve IPs is branch-and-bound (B\&B) \cite{land2010automatic}, which lies at the core of common solvers such as SCIP, Gurobi, and CPLEX. 
Thus, most of existing methods improve the performance of a solver on a distribution of instances, by training models for critical search decisions in B\&B such as variable and node selection \cite{he2014learning,khalil2017learn,gasse2019exact}. Nevertheless, these methods are generally limited to small instances and require sufficient interface access to the internal solving process of the solvers.



This paper mainly tackle the issue that \emph{how to improve a solver from externals such that it can find high-quality solutions more quickly?} In specific, we propose a high-level, learning based LNS method to solve general IP problems. Based on deep reinforcement learning (RL), we train a policy network as the destroy operator in LNS, which decides a subset of variables in the current solution for reoptimization. Then we use a solver as the repair operator, which solves sub-IPs to reoptimize the destroyed variables. Despite being heuristic, our method can effectively handle the large-scale IP by solving a series of smaller sub-IPs. Moreover, complex interface to the solver's internal logic is not required. However, the above RL task is challenging, mainly because the action space, i.e., number of variable subsets at each LNS step, is exponentially large. To resolve this issue, we represent all the subsets by factorizing them into binary decisions on each variable, i.e., whether a variable should be destroyed. In doing so, we make it possible to learn a policy to select \emph{any subset} from large discrete action spaces (at least $2^{1000}$ candidates in our experiments). To this end, we design a Graph Neural Network (GNN) based policy network that enables learning policies for each variable in parallel, and train it by a customized actor-critic algorithm.

A recent work \cite{song2020general} also attempts to learn LNS policy to solve IP problems, and we generalize this framework to enable learning more flexible and powerful LNS algorithms. One limitation of \cite{song2020general} is that it hypothesizes a constant cardinality of the destroyed variable subset at each LNS step, which is a predefined hyperparameter. However, the number and choice of optimized variables at each step should be adaptive according to instance information and solving status, which is achieved in our method.
In doing so, the LNS policies trained by our method significantly outperform those trained by the method in \cite{song2020general}.




We evaluate our method on four NP-hard benchmark problems with SCIP as the repair solver. Extensive results show that our method generally delivers better solutions than SCIP with mere 1/2 or 1/5 of runtime, and significantly outperforms LNS baselines with the same runtime. These advantages notably persist when the trained policies are directly applied to much larger problems. We also apply our LNS framework to Gurobi, which shows superiority over the solver itself and other baselines.


\section{Related work}
\label{related_work}

In this section, we briefly review existing works related to ours. We first describe two main streams of learning based methods to solve COPs, and then introduce the literature that study RL with large discrete action space, which is also an essential issue we confront in this paper.



\paragraph{Learning to solve specific COPs.} 
Quite a few works attempt to learn heuristics to solve certain types of COPs. Compared to construction ones, methods that learn improvement heuristics can often deliver smaller optimality gap, by training policies to iteratively improve the solution. \citet{chen2019learning} propose to learn how to locally rewrite a solution; \citet{wu2019learning} train policies to pick the next solution in 
local moves; \citet{lu2019learning} learn to select local operators to reform a solution. These methods are generally limited by simple local operations. A few methods learn more powerful operators under the LNS framework. \citet{hottung2019neural} train an attention model to repair the solution every time it is broken by a predefined destroy operator. Similarly, \citet{gao2020learn} combine GNN and Recurrent Neural Network to learn a reinsertion operator to repair sequence-based solution. However, all the above methods are limited to specific problem types, e.g., the LNS methods in \cite{hottung2019neural,gao2020learn} are designed only for routing problems. 
In contrast, this paper aims to solve general IP problems with a high-level LNS framework and raise its performance by learning better policies.


\paragraph{Learning to solve IP problems.} Most of learning based methods for IPs aim to improve inner policies of B\&B algorithm. For example, \citet{he2014learning} learn to explore nodes in B\&B tree by imitating an oracle. \citet{gasse2019exact} train a GNN model to predict the strong branching rule by imitation learning. \citet{khalil2017learn} predict the use of primal heuristics by logistic regression. Other components of B\&B are also considered to improve its practical performance, such as learning to select cutting planes by \citet{tang2020reinforcement}, predicting local branching at root node by \citet{ding2020accelerating} or refining the primal heuristic and branching rule concurrently by \citet{nair2020solving}. Different from these works, we employ RL to improve practical performance of IP solvers especially for large-scale problems, without much engineering effort on interfacing with inner process of solvers.\footnote{Nevertheless, our method can also work with solvers enhanced by the above methods as repair operators.} This is also noted in \cite{song2020general}, which proposes to combine learning and LNS to solve IPs. However, a major drawback of this method is that the subsets of destroyed variables are assumed to be fixed-sized, which limits its performance. In contrast, our LNS framework allows picking variable subsets in a more adaptive way, and thus empowers learning broader classes of policies.

\paragraph{RL with large action spaces.} Learning with large-sized, high-dimensional discrete action spaces is still intricate in current RL research. In this direction, \citet{pazis2011generalized} use binary format to encode all actions, and learn value functions for each bit. \citet{tavakoli2018action} design neural networks with branches to decide on each component of the action. Besides, \citet{dulac2015deep} and \citet{chandak2019learning} update the action representation by solving continuous control problems so as to alleviate learning complexity via generalization. Other similar works can be found in \cite{tang2020discretizing,hubert2021learning,andrychowicz2020learning}. However, we note that all these methods are generally designed for tasks with (discretized) continuous action spaces, which are relatively low-dimensional and small-sized (at most $10^{25}$ as in \cite{tavakoli2018action}). In contrast, action spaces in our scenario are much larger (at least $2^{1000}$). In this paper, we propose to factorize the action space into binary actions on each dimension, and train the individual policies in parallel through parameter sharing.

\section{Preliminaries}
\textbf{Integer Program (IP)} is a typically defined as $\arg\min_{x}\{\mu^{\top}x|Ax\leq b; x\geq0; x\in \mathbb{Z}^{n}\}$, where $x$ is a vector of $n$ decision variables; $\mu\in\mathbb{R}^n$ denotes the vector of objective coefficients; the incidence matrix $A\in\mathbb{R}^{m\times n}$ and right-hand-side (RHS) vector $b\in \mathbb{R}^{m}$ together define $m$ linear constraints. With the above formulation, the size of an IP problem is generally reflected by the number of variables ($n$) and constraints ($m$).

\textbf{Large Neighborhood Search (LNS)} is a type of improvement heuristics, which iteratively reoptimizes a solution by the \emph{destroy} and \emph{repair} operator until certain termination condition is reached \cite{pisinger2010large}. Specifically, the former one breaks part of the solution $x_t$ at step $t$, then the latter one fixes the broken solution to derive the next solution $x_{t+1}$. The destroy and repair operator together define the solution neighborhood $\mathcal{N}(x_{t})$, i.e., solution candidates that can be accessed at time step $t$. Compared to typical local search heuristics, LNS is more effective in exploring the solution space, since a larger neighborhood is considered at each step \cite{pisinger2010large}. Most of existing LNS methods rely on problem specific operators, e.g., removal and reinsertion for solutions with sequential structure \cite{ropke2006adaptive,prescott2009branch}. Instead, we propose a RL based LNS for general IP problems, with the learned destroy operator to select variables for reoptimization at each step.

\section{Methodology}
\label{sec:method}
In this section, we first formulate our LNS framework as a Markov Decision Process (MDP). Then, we present the factorized representation of the large-scale action space, and parametrize the policy by a specialized GNN. Finally, we introduce a customized actor-critic algorithm to train our policy network for deciding the variable subset.

\subsection{MDP formulation}
Most of existing works learn LNS for specific problems \cite{hottung2019neural,gao2020learn}, which rely on extra domain knowledge and hinder their applicability to other COPs. In this paper, we apply LNS to general IP problems, in which RL is employed to learn a policy that at each step, selects a variable subset from the solution to be reoptimized. We formulate this sequential decision problem as a discrete-time MDP. Specifically, we regard the destroy operator as the agent and the remainder in LNS as the environment. In one episode of solving an IP instance, the MDP can be described as follows:

\noindent\textbf{States.} A state $s_t\in \mathcal{S}$ at each LNS step needs to not only reflect the instance information but also the dynamic solving status. For the former, we represent it by static features of variables and constraints, along with the incidence matrix in an instance. The latter is represented by dynamic features, including the current solution $x_t$ and dynamic statistics of incumbent solutions up to step $t$. Specifically, both the current and incumbent solution at $t=0$ are defined as the initial solution $x_0$.

\noindent\textbf{Actions.} At each state $s_t$, an action of the agent is to select a variable subset $a_t$ from all candidate subsets $\mathcal{A}$ for reoptimization.

\noindent\textbf{Transition.} The repair operator (i.e., an IP solver) solves the sub-IP, where only the variables in action $a_t$ are optimized with the others equaling to their current values in $x_t$. Accordingly, the next state $s_{t+1}$ is deterministically attained by updating $s_{t}$ with the new solution: 
\begin{equation} \label{eq:new_solution}
x_{t+1}=\arg\min_{x}\{\mu^{\top}x|Ax\leq b; x\geq0; x\in \mathbb{Z}^{n}; x^i=x_t^i, \forall x^i\notin a^t\}.
\end{equation}
\noindent\textbf{Rewards.} The reward function is defined as $r_t=r(s_t,a_t)=\mu^{\top}(x_t-x_{t+1})$, which is the change of the objective value. Given the step limit $T$ of the interaction between the agent and environment, the return (i.e., cumulative rewards) from step $t$ of the episode is $R_t=\sum_{k=t}^{T} \gamma^{k-t} r_k$, with the discount factor $\gamma\in [0,1]$.  The goal of RL is to maximize the expected return $\mathbb{E}[R_1]$ over all episodes, i.e., the expected improvement over initial solutions, by learning a policy.

\noindent\textbf{Policy.} The stochastic policy $\pi$ represents a conditional probability distribution over all possible variable subsets given a state. Starting from $s_0$, it iteratively picks an action $a_t$ based on the sate $s_t$ at each step $t$, until reaching the step limit $T$.

\begin{figure}[t!]
   \centering
   \includegraphics[width=1.48\linewidth]{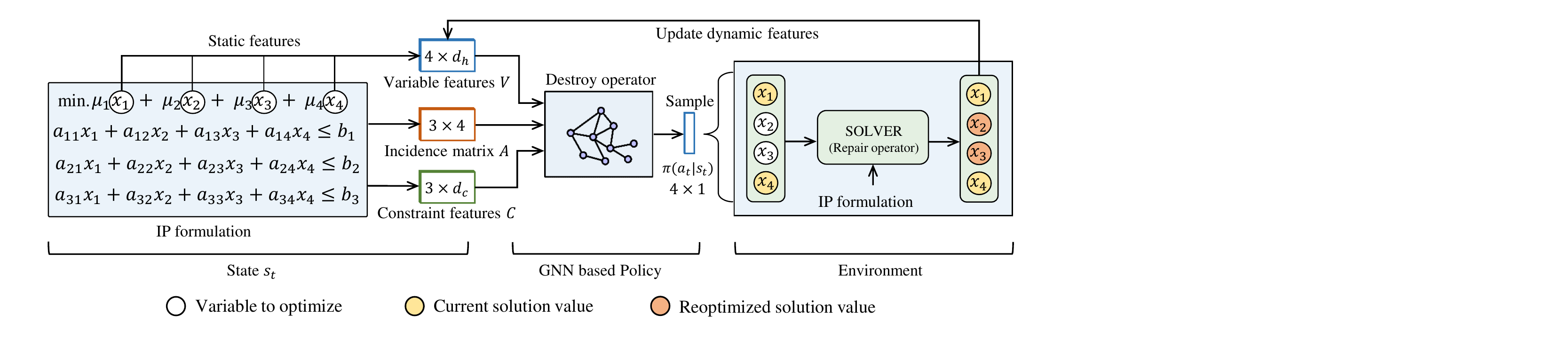}
   \vspace{-0.5cm}
   \caption{\textbf{An example of LNS solving an instance with 4 variables and 3 constraints.} Given the current state characterized by static and dynamic features, the RL agent (destroy operator) selects a subset of variables from the current solution to be reoptimized. Then this action influences the environment, through which a sub-IP is formulated and solved by the repair operator (i.e., solver). As feedbacks, the new solution updates the state and a reward is provided to the agent. The above process is repeated until reaching the time limit.}
    \label{fig:network}
\end{figure}

An example of the proposed LNS framework solving an IP instance is illustrated in Figure~\ref{fig:network}, in which the GNN based policy network has the potential to process instances of any size via sharing parameters across all variables.


\subsection{Action factorization} \label{sec:factorization}

Given the vector of decision variables $x$ in an IP instance, we gather all its elements in a set $X=\{x^1,x^2,\ldots,x^n\}$. Accordingly, we define the action space $\mathcal{A}=\{a|a\subseteq
X\}$, which contains all possible variable subsets. Thus our RL task is learning a policy to select $a_t\in \mathcal{A}$ for reoptimization at each step $t$ of LNS, and the cardinality of $a_t$ (i.e., $|a_t|$) reflects the destroy degree on the solution. Apparently, the size of the combinatorial space $\mathcal{A}$ is $2^n$, which grows exponentially with the number of variables $n$. This prevents the application of general RL algorithms on large-scale problems, since they require an explicit representation of all actions and exploration over such huge action space. 

As a special case under our LNS framework, \citet{song2020general} assumes the action space as the subspace of $\mathcal{A}$ that merely contains equal-sized variable subsets, that is, $\mathcal{A}_z=\{a_z|a_z\in\mathcal{A},|a_z|=z\}$. In doing so, they instead learn classifying variables into groups of equal size, to optimize each iteratively. Despite the good performance on some COPs, such action representation with the fixed destroy degree makes the LNS search inflexible and thus limits the class of policies that can be learned. Also, the issue of large action space still exists, since the patterns to 
group variables could be combinatorial. 

To involve larger action space and in the meanwhile keep the learning tractable, we factorize the combinatorial action space $\mathcal{A}$ into elementary actions on each dimension (i.e., variable). Specifically, we denote $a_t^i\in \{x^i, \emptyset\}$ as the elementary action for variable $x^i$ at step $t$. It means that $x^i$ is either selected for reoptimization, or not selected and thus fixed as its value in the current solution. With such representation, 
any variable subset can also be expressed as $a_t=\bigcup_{i=1}^{n}a_t^i$.
Therefore, our task can be converted into learning policies for binary decisions on each variable. This enables exploration of large action space by traversing binary action spaces, the number of which grows linearly with the number of variables. To this end, we factorize the original policy as below:
\begin{equation}   \label{eq:union_policy}
\pi(a_t|s_t)={\prod}_{i=1}^{n}\pi^i(a_t^i|s_t),
\end{equation}
which expresses $\pi(a_t|s_t)$, the probability of selecting an action, as the product of probabilities of selecting its elements. Since ${\sum}_{a_t^i\in\{x^i,\emptyset\}}\pi^i(a_t^i|s_t)=1,\,\forall\, i\in \{1,\ldots,n\}$, it is easy to verify that the sum of probabilities of all actions in $\mathcal{A}$ still equals to $1$, i.e., ${\sum}_{a_t\in\mathcal{A}}\pi(a_t|s_t)=1$. Based on this factorization, we can apply RL algorithms to train policies $\pi^i$ for each variable $x^i$. In the following subsections, we first parametrize the policy $\pi$ by a GNN $\pi_\theta$, and then train it by a customized actor-critic algorithm. 

\subsection{Policy parametrization}
\label{sec:policy}

Policy networks in deep RL algorithms are generally designed to map states to probabilities of selecting each action. In our case, the complexity of directly training such policy networks could exponentially increase with the growing number of variables. Based on Equation~(\ref{eq:union_policy}), an ad hoc way to reduce the complexity is training individual policies for each dimension. However, the IP problems in our study comprise a high volume of variables, and it is unmanageable to train and store this many networks. Learning disjoint policies without coordination could also suffer from convergence problem as shown in \cite{matignon2012independent}. Another possible paradigm could be constructing a semi-shared network, in which all dimensions share a base network and derive outputs from separate sub-networks. Following a similar idea, \citet{tavakoli2018action} design a network with branches for deep Q-learning algorithms. However, it is not suitable for IP problems, since the number of sub-networks is only predefined for single problem size, without the generalization ability to different-sized instances.

To circumvent the above issues, we design a GNN based policy network to share parameters across all dimensions, and also enable generalization to any problem size \cite{battaglia2018relational,wu2020comprehensive}. To this end, we first describe the state $s_t$ by a bipartite graph $\mathcal{G}=(\mathcal{V},\mathcal{C},\mathbf{A})$, where $\mathcal{V}=\{v_1,\cdots,v_n\}$ denotes variable nodes with features $\mathbf{V}\in\mathbb{R}^{n\times d_v}$; $\mathcal{C}=\{c_1,\cdots,c_m\}$ denotes constraint nodes with features $\mathbf{C}\in\mathbb{R}^{m\times d_c}$; $\mathbf{A}\in\mathbb{R}^{m\times n}$ denotes the adjacency matrix with $a_{ji}$ being the weight (or feature) of the edge between nodes $c_j$ and $v_i$, which is practically the incidence matrix $A$ in an IP instance. This state representation is similar to the one in \cite{gasse2019exact}, which depicts the status in B\&B tree search to learn branching policy. We extend its usage outside the solver to reflect the solving process in our LNS framework.

Then, we parametrize the policy as $\pi_\theta(a_t|s_t)$ by a graph convolutional network (GCN), a GNN variant broadly used in various tasks \cite{kipf2017semi,duvenaud2015convolutional,yao2019graph}. The architecture of our design is illustrated in Appendix A.1, with the graph convolution layer expressed as below:
\begin{equation}  
\begin{aligned}
&\mathbf{C}^{(k+1)}=\mathbf{C}^{(k)}+\sigma\left(\text{LN}\left(\mathbf{A}\mathbf{V}^{(k)}W_v^{(k)}\right)\right), \\
&\mathbf{V}^{(k+1)}=\mathbf{V}^{(k)}+\sigma\left(\text{LN}\left(\mathbf{A}^{\top}\mathbf{C}^{(k+1)}W_c^{(k)}\right)\right), k=0,\ldots,K\\
\end{aligned}
\end{equation}
where $W_v^{(k)}, W_c^{(k)}\in \mathbb{R}^{d_h\times d_h}$ are trainable weight matrices in the $k$-th layer; $\mathbf{V}^{(k)}=[\mathbf{v}_1^{(k)}\cdots \mathbf{v}_n^{(k)}]^{\top}$ and $\mathbf{C}^{(k)}=[\mathbf{c}_1^{(k)}\cdots \mathbf{c}_m^{(k)}]^{\top}$ are node embeddings for variables and constraints respectively in $k$-th layer; LN and $\sigma(\cdot)$ denote layer normalization and Tanh activation function respectively. In particular, we linearly project raw features of variables and constraints into the initial node embeddings $\mathbf{V}^{(0)}$ and $\mathbf{C}^{(0)}$ with $d_h$ dimensions ($d_h=128$), and keep this dimension through all layers. After $K$ iterations of convolution ($K=2$), the embeddings for the two clusters of heterogeneous nodes are advanced as $\{\mathbf{v}_i^{K}\}_{i=1}^n$ and $\{\mathbf{c}_j^{K}\}_{j=1}^m$. We finally process the former by a multi-layer perceptron (MLP) with a single-value output activated by Sigmoid. In this way, the output value can represent the probability of a variable being selected, i.e., $\pi^i(a_t^i|s_t)=\pi_\theta(a_t^i|s_t)=\text{MLP}(\mathbf{v}_i^{K}), \forall\, i\in \{1,\ldots,n\}$, such that we can conduct Bernoulli sampling accordingly on each variable and attain the subset. In this paper, we structure the MLP by two hidden layers, which have 256 and 128 dimensions respectively and are activated by Tanh. 

\subsection{Training algorithm} 
Actor-critic is one of the policy gradient methods developed from REINFORCE \cite{williams1992simple}. It parametrizes state-value or action-value function as a critic network to estimate the expected return. In this paper, we adopt Q-actor-critic with $Q_\omega(s,a) \approx Q(s,a) = \mathbb{E}[R_t|s_t=s,a_t=a]$, where $\omega$ is the parameter set to be learned. In doing so, $Q_\omega$ (i.e., the critic) can be updated through bootstrapping and leveraged in the training of the policy network (i.e., the actor). Specifically, the loss functions for the critic and the actor are defined as follows: 
\begin{align}
&L(\omega)=\mathbb{E}_{\mathcal{D}}[(\gamma Q_\omega(s_{t+1},a_{t+1})+r_t-Q_\omega(s_t,a_t))^2], \label{eq:critic} \\[5pt]
&L(\theta)=\mathbb{E}_{\mathcal{D}}[Q_\omega(s_t,a_t) \log\pi_\theta(a_t|s_t)], \label{eq:actor} 
\end{align}
where the experience replay buffer $\mathcal{D}$ contains transitions $(s_t,a_t,r_t,s_{t+1},a_{t+1})$, which are collected during solving a batch of instances. $\gamma Q_\omega(s_{t+1},a_{t+1})+r_t$ is the one-step temporal-difference (TD) target to update the critic.



The above Q-actor-critic is not directly applicable to the primitive high-dimensional action spaces in our IP problems. To learn the factorized policies we designed in Section \ref{sec:factorization}, one way is to customize it by training an actor and critic on each dimension, and updating the parameters via the loss functions:
\begin{align}
&\tilde{L}(\omega)=\mathbb{E}_{\mathcal{D}}[\frac{1}{n}{\sum}_{i=1}^{n}(\gamma Q_\omega(s_{t+1},a_{t+1}^i)+r_t-Q_\omega(s_t,a_t^i))^2], \label{eq:critic_new} \\
&\tilde{L}(\theta)=\mathbb{E}_{\mathcal{D}}[\frac{1}{n}{\sum}_{i=1}^{n}Q_\omega(s_t,a_t^i)\log(\pi_{\theta}(a_t^i|s_t))], \label{eq:actor_new} 
\end{align}

where the parameter-sharing $Q_\omega$ is used across all dimensions as the actor $\pi_\theta$, i.e., $Q_\omega(s_t,a_t^i)=Q^i(s_t,a_t^i), \forall\, i\in \{1,\ldots,n\}$. However, our experiments show that the critic trained by bootstrapping on each elementary action delivers inferior performance, which is similar to the finding in \cite{tavakoli2018action}. Intuitively, this may stem from excessive action-value regressions with one single neural network. To circumvent this issue, we keep the global TD learning in Equation~(\ref{eq:critic}) and adjust elementary policies by the $Q$ value of the state-action pair $(s_t,a_t)$, such that:
\begin{equation} \label{eq:actor_final}
\tilde{L}(\theta)=\mathbb{E}_{\mathcal{D}}[Q_\omega(s_t,a_t)  {\sum}_{i=1}^{n}\log(\pi_{\theta}(a_t^i|s_t))],
\end{equation}
where the critic $Q_\omega$ is structured in the same manner as $\pi_{\theta}$, except that: 1) we add a binary value to the raw features of each variable $a_t^i$ to indicate whether it is selected; 2) we use MLP to process the graph embedding, which aggregates the embeddings of variables by mean-pooling, to output a real value that represents the $Q$ value.


\noindent\textbf{Clipping \& masking.} To enhance exploration, we clip the probabilities of being selected for each variable in a range $[\epsilon,1-\epsilon]$, $\epsilon<$0.5. It helps avoid always or never traversing some variables with extreme probabilities. We also do not consider empty or universal sets of variables, which lead to unsuitable sub-IPs. Though the chance to select these two sets are low, we mask them by resampling. 

Details of the training algorithm are given in Appendix A.2. Besides the policy network we designed in Section \ref{sec:policy}, we also adopt this algorithm to train a MLP based semi-shared network similar to the one in \cite{tavakoli2018action}, which indicates that the fully-shared one (ours) is more suitable for IP problems. More details are given in Appendix A.3.


\section{Experimental results} \label{sec:experiments}
We perform experiments in this section on four NP-hard benchmark problems: Set Covering (SC), Maximal Independent Set (MIS), Combinatorial Auction (CA) and  Maximum Cut (MC), which are widely used in existing works. Our code is available.\footnote{https://github.com/WXY1427/Learn-LNS-policy}

\noindent\textbf{Instance generation.}
We generate SC instances with $1000$ columns and $5000$ rows following the procedure in \cite{balas1980set}. MIS instances are generated following \cite{bergman2016decision}, where we use the Erd\H{o}s-R\'{e}nyi random graphs with $1500$ nodes and set the affinity number to $4$. CA instances with $2000$ items and $4000$ bids are generated according to arbitrary relationships in \cite{leyton2000towards}. MC instances are generated according to Barabasi-Albert random graph models \cite{albert2002statistical}, with average degree $4$, and we adopt graphs of $500$ nodes. For each problem type, we generate $100$, $20$, $50$ instances for training, validation, and testing. In addition, we double and quadruple the number of variables and generate $50$ larger and even larger instances respectively for each problem, to verify the generalization performance. We name the instance groups and display their average sizes in Table \ref{tab:pro_size}.

\begin{table}[hbt!]  \small
	\centering
	\caption{Average sizes of problem instances.}
	\begin{threeparttable}
    \scalebox{0.92}{	
	\begin{tabular}{crrrrrrrrrrrr}
		\toprule
		 & \multicolumn{4}{c}{Training} & \multicolumn{8}{c}{Generalization} \\
		\cmidrule(lr){2-5}
		\cmidrule(lr){6-13}
		Num. of & SC & MIS & CA & MC  & SC$_{2}$ & MIS$_{2}$ & CA$_{2}$ & MC$_{2}$ & SC$_{4}$ & MIS$_{4}$ & CA$_{4}$ & MC$_{4}$ \\ \midrule
		Variables   & 1000  & 1500 & 4000 & 2975 & 2000 & 3000 & 8000 & 5975 & 4000 & 6000 & 16000 & 11975  \\
		Constraints   & 5000 & 5939 & 2674 & 4950 & 5000 & 11932 & 5357 & 9950 & 5000 & 23917 & 10699 & 19950 \\	
		\bottomrule          
	\end{tabular}}
	\end{threeparttable}
	\label{tab:pro_size}
\end{table}

\noindent\textbf{Features.} To represent an instance, we extract static features of variables and constraints, along with the incidence matrix after presolving by the solver at the root node of B\&B. In this way, redundant information could be removed and the extracted features could be more clear and compact in reflecting the problem structure. For the dynamic features, we record for each variable its value in the current solution and incumbent, as well as its average value in all incumbents to represent the solving status of LNS. Specially, at the step $t=0$, the average incumbent values are naturally attained from the initial solution, i.e., the incumbent at the root node. Note that in practice, we concatenate the static and dynamic features for each variable, and attach them ($\mathbf{V}$) to variable nodes in the bipartite graph. The features of constraints ($\mathbf{C}$) and the incidence matrix ($A$) are attached to the constraint nodes and edges, respectively. Detailed description of the above features are available in Appendix A.4.


\noindent\textbf{Hyperparameters.} We use the state-of-the-art open source IP solver SCIP (v6.0.1) \cite{achterberg2009scip} as the repair operator, which also serves as a major baseline. We run all experiments on an Intel(R) Xeon(R) E5-2698 v4 2.20GHz CPU. For each problem, we train 200 iterations, during each we randomly draw $\mathit{M}$=$10$ instances. We set the training step limit $T$=$50$, $50$, $70$, $100$ for SC, MIS, CA and MC respectively. The time limit for repair at each step is 2 seconds, unless stated otherwise. We use $\epsilon$=0.2 for probability clipping. For the Q-actor-critic algorithm, we set the length of the experience replay $TM$, the number of updating the network $U$=$4$ and the batch size $\mathcal{B}$=$TM/U$. We set the discount factor $\gamma$=0.99 for all problems, and use Adam optimizer with learning rate $1\times 10^{-4}$. 
To show the applicability of our method on other solvers, we have also performed experiments where Gurobi \cite{gurobi} is used as the repair solver, which will be discussed in Section \ref{sec:gurobi}.

\subsection{Comparative analysis} \label{sec:exp_one}

\noindent\textbf{Baselines.} We compare our method with four baselines: 
\begin{itemize}[leftmargin=*]
    \item SCIP with default settings. 
    \item U-LNS: a LNS version which uniformly samples a subset size, then fills it by uniformly sampling variables. We compare with it to show that our method can learn useful subset selection policies.
    \item R-LNS: a LNS version with hand-crafted rule proposed in \cite{song2020general}, which randomly groups variables into disjoint equal-sized subsets and reoptimizes them in order.
    \item FT-LNS: the best-performing LNS version in \cite{song2020general}, which applies forward training, an imitation learning algorithm, to mimic the best demonstrations collected from multiple R-LNS runs.
\end{itemize}


Following \cite{song2020general}, we tune the group number of R-LNS (and FT-LNS since it imitates R-LNS) from $2$ to $5$, and apply the best one to each problem. To train FT-LNS, we collect $10$ demonstrations for each instance, and tune the step limit to 20 for SC, MIS, CA and 50 for MC, which perform the best. Same as our method, all LNS baselines also use SCIP as the repair operator with 2s time limit. To compare solution quality, we use the average objective value and standard deviation over the 50 testing instances as metrics. Also, since all problems are too large to be solved optimally, we measure the primal gap \cite{khalil2017learn} to reflect the difference between the solution $\tilde{x}$ of a method to the best one $x^{\ast}$ found by all methods. We compute  $|\mu^{\top}\tilde{x}-\mu^{\top}x^{\ast}|/max\{|\mu^{\top}\tilde{x}|,|\mu^{\top}x^{\ast}|\}\cdot100\%$ for each instance, then average the gaps for all 50 ones. Below we report the results on testing instances of the same sizes as in training. 

\begin{table}[!t]  \small
	\centering
	\caption{Comparison with SCIP and LNS baselines.}
	\begin{threeparttable}
    \scalebox{0.86}{
	\begin{tabular}{ccccccccccccc}
		\toprule
	    & \multicolumn{3}{c}{SC} & \multicolumn{3}{c}{MIS}   & \multicolumn{3}{c}{CA}    & \multicolumn{3}{c}{MC}\\
		\cmidrule(lr){2-4}
		\cmidrule(lr){5-7}\cmidrule(lr){8-10}\cmidrule(lr){11-13}
			Methods & \multicolumn{2}{c}{Obj.$\pm$Std.\%} & Gap\% & \multicolumn{2}{c}{Obj.$\pm$Std.\%} &Gap\% & \multicolumn{2}{c}{Obj.$\pm$Std.\%} & Gap\%&\multicolumn{2}{c}{Obj.$\pm$Std.\%} & Gap\%\\ \midrule
			SCIP   & \multicolumn{2}{c}{567.66 $\pm$ 8.76} & 3.62 & \multicolumn{2}{c}{-681.02 $\pm$ 1.14} &0.29 & \multicolumn{2}{c}{-110181 $\pm$ 2.03} & 2.98 & \multicolumn{2}{c}{-852.57 $\pm$ 1.22} & 4.37 \\
			$\text{SCIP}^{*}$  & \multicolumn{2}{c}{552.82 $\pm$ 8.69} & 0.91 & \multicolumn{2}{c}{-681.76 $\pm$ 1.06} & 0.18 & \multicolumn{2}{c}{-111511 $\pm$ 1.85} & 1.85 &\multicolumn{2}{c}{-861.10 $\pm$ 1.26} &3.41 \\ 
			$\text{SCIP}^{**}$  & \multicolumn{2}{c}{\textbf{550.68 $\pm$ 8.60}} & \textbf{0.53} & \multicolumn{2}{c}{-682.46 $\pm$ 1.02} & 0.07 & \multicolumn{2}{c}{-112638 $\pm$ 1.68} & 0.82 &\multicolumn{2}{c}{-863.63 $\pm$ 1.32} & 3.13 \\ 
			U-LNS & \multicolumn{2}{c}{568.60 $\pm$ 12.17} & 5.99 & \multicolumn{2}{c}{-681.38 $\pm$ 0.95} & 0.23  & \multicolumn{2}{c}{-103717 $\pm$ 1.92} & 8.67 &\multicolumn{2}{c}{-869.20 $\pm$ 1.53} & 2.50 \\ 	
			R-LNS  &\multicolumn{2}{c}{560.54 $\pm$ 8.07}  & 2.38 & \multicolumn{2}{c}{-682.20 $\pm$ 0.93}  & 0.11 & \multicolumn{2}{c}{-109550 $\pm$ 1.62} & 3.44 &\multicolumn{2}{c}{-882.18 $\pm$ 1.27} & 1.05 \\ 
			FT-LNS  & \multicolumn{2}{c}{564.00 $\pm$ 8.03} & 3.02 & \multicolumn{2}{c}{-681.82 $\pm$ 0.93} & 0.17  & \multicolumn{2}{c}{-107370 $\pm$ 2.03} & 5.45 & \multicolumn{2}{c}{-867.05 $\pm$ 1.64} & 2.75 \\ 	
			Ours  &\multicolumn{2}{c}{551.50 $\pm$ 8.59}  & 0.68 & \multicolumn{2}{c}{\textbf{-682.52 $\pm$ 0.98}}  & \textbf{0.06} & \multicolumn{2}{c}{\textbf{-112666 $\pm$ 1.72}} & \textbf{0.77} & \multicolumn{2}{c}{\textbf{-889.61 $\pm$ 1.32}} & \textbf{0.27} \\ 	
		\bottomrule          
	\end{tabular}}
	\begin{tablenotes} \small
		\item[\textbf{1}] $*$ and $**$ mean the method run with $500$s and $1000$s.
	\end{tablenotes}	
	\end{threeparttable}
	\label{tab:long}
\end{table}


In this paper, we aim to improve an IP solver from externals to enable more efficient search of high-quality solutions in a broad range of COPs. To this end, we compare all methods for time-bounded optimization with the same 200s time limit, and further allow SCIP to run for longer time, i.e., 500s and 1000s. The results are gathered in Table \ref{tab:long}. As shown, our method significantly outperforms all baselines on all problems with the same 200s time limit. It is notable that FT-LNS is inferior to R-LNS which yields demos for its imitation learning. The reason might be that FT-LNS only mimics the random demos of short (training) step limits, and hence lacks the ability of generalizing to longer steps. This limitation might hinder its application since in practice, IP problems are often solved in an \emph{anytime} manner with flexible time/step limits. In contrast, our method avoids this myopic issue by RL training.
In Appendix A.5, we also show that FT-LNS can outperform R-LNS with the same number of LNS steps as in training.
Another key observation from Table \ref{tab:long} is that with longer time limits, SCIP is able to find better solutions than the three LNS baselines on SC, MIS and CA. However, our method still surpasses SCIP (500s) on all problems and SCIP (1000s) on MIS, CA and MC.


\subsection{Generalization analysis}

\begin{table}[!t]  \small
\setlength{\tabcolsep}{5pt}
	\centering
	\caption{Generalization to large instances.}
	\begin{threeparttable}
    \scalebox{0.88}{	
	\begin{tabular}{ccccccccccccc}
		\toprule
		& \multicolumn{3}{c}{SC$_{2}$} & \multicolumn{3}{c}{MIS$_{2}$}   & \multicolumn{3}{c}{CA$_{2}$}& \multicolumn{3}{c}{MC$_{2}$}    \\
		\cmidrule(lr){2-4}
		\cmidrule(lr){5-7}\cmidrule(lr){8-10} \cmidrule(lr){11-13}
			Methods & \multicolumn{2}{c}{Obj.$\pm$Std.\%} & Gap\% & \multicolumn{2}{c}{Obj.$\pm$Std.\%} &Gap\% & \multicolumn{2}{c}{Obj.$\pm$Std.\%} & Gap\% & \multicolumn{2}{c}{Obj.$\pm$Std.\%} & Gap\%   \\ \midrule
			SCIP   & \multicolumn{2}{c}{303.18 $\pm$ 8.62} & 6.80 & \multicolumn{2}{c}{-1323.90 $\pm$ 0.81} & 3.22 & \multicolumn{2}{c}{-205542 $\pm$ 2.87} & 7.14 &  \multicolumn{2}{c}{-1691.48 $\pm$ 1.18} & 6.22\\
			$\text{SCIP}^{*}$  & \multicolumn{2}{c}{298.12 $\pm$ 8.08} & 5.04 & \multicolumn{2}{c}{-1357.04 $\pm$ 1.34} & 0.80 & \multicolumn{2}{c}{-214654 $\pm$ 1.44} & 3.02 & \multicolumn{2}{c}{-1706.45 $\pm$ 1.26} & 5.39 \\ 
			$\text{SCIP}^{**}$  & \multicolumn{2}{c}{\textbf{295.70 $\pm$ 7.89}} & \textbf{4.21} & \multicolumn{2}{c}{-1361.98 $\pm$ 1.06} & 0.44 & \multicolumn{2}{c}{\textbf{-217271 $\pm$ 1.93}} & \textbf{1.84} & \multicolumn{2}{c}{-1714.71 $\pm$ 1.02} & 4.93 \\ 
			U-LNS & \multicolumn{2}{c}{303.36 $\pm$ 8.24} & 6.90 & \multicolumn{2}{c}{-1364.66 $\pm$ 0.69} & 0.24  & \multicolumn{2}{c}{-197453 $\pm$ 1.86} & 10.79 & \multicolumn{2}{c}{-1769.00 $\pm$ 1.03} & 1.92 \\ 	
			R-LNS  &\multicolumn{2}{c}{300.84 $\pm$ 7.74}  & 6.06 & \multicolumn{2}{c}{-1339.00 $\pm$ 0.81}  & 2.12 & \multicolumn{2}{c}{-204145 $\pm$ 1.57} & 7.77 & \multicolumn{2}{c}{-1767.09 $\pm$ 1.00} & 2.03\\ 
			FT-LNS  & \multicolumn{2}{c}{303.52 $\pm$ 7.86} & 6.98 & \multicolumn{2}{c}{-1345.58 $\pm$ 0.86} & 1.63  & \multicolumn{2}{c}{-212264 $\pm$ 1.35} & 4.10 & \multicolumn{2}{c}{-1700.58 $\pm$ 1.64} & 5.72\\ 	
			Ours  &\multicolumn{2}{c}{297.90 $\pm$ 8.20}  & 4.98 & \multicolumn{2}{c}{\textbf{-1367.78 $\pm$ 0.68}}  & \textbf{0.01} & \multicolumn{2}{c}{-216006 $\pm$ 1.15} & 2.40 & \multicolumn{2}{c}{\textbf{-1803.71 $\pm$ 0.92}} & \textbf{0.00}\\ 	\midrule
		& \multicolumn{3}{c}{SC$_{4}$} & \multicolumn{3}{c}{MIS$_{4}$}   & \multicolumn{3}{c}{CA$_{4}$}& \multicolumn{3}{c}{MC$_{4}$}    \\
		\cmidrule(lr){2-4}
		\cmidrule(lr){5-7}\cmidrule(lr){8-10} \cmidrule(lr){11-13}
			Methods & \multicolumn{2}{c}{Obj.$\pm$Std.\%} & Gap\% & \multicolumn{2}{c}{Obj.$\pm$Std.\%} &Gap\% & \multicolumn{2}{c}{Obj.$\pm$Std.\%} & Gap\% & \multicolumn{2}{c}{Obj.$\pm$Std.\%} & Gap\%   \\ \midrule			
			SCIP & \multicolumn{2}{c}{179.88 $\pm$ 6.35} & 6.37 & \multicolumn{2}{c}{-2652.56 $\pm$ 0.58} &3.05 & \multicolumn{2}{c}{-372291 $\pm$ 1.22} & 13.44 & \multicolumn{2}{c}{-3392.02 $\pm$ 0.86} & 5.45\\
			$\text{SCIP}^{*}$  &  \multicolumn{2}{c}{177.44 $\pm$ 6.64} & 4.91 & \multicolumn{2}{c}{-2652.56 $\pm$ 0.58} &3.05 & \multicolumn{2}{c}{-372291 $\pm$ 1.22} & 13.44 & \multicolumn{2}{c}{-3392.94 $\pm$ 0.82} & 5.43\\ 
			$\text{SCIP}^{**}$  & \multicolumn{2}{c}{\textbf{175.38 $\pm$ 6.98}} & \textbf{3.68} & \multicolumn{2}{c}{-2673.64 $\pm$ 1.54} &2.28 & \multicolumn{2}{c}{-372291 $\pm$ 1.22} & 13.44 &\multicolumn{2}{c}{-3394.42 $\pm$ 0.81} & 5.39\\ 
			U-LNS & \multicolumn{2}{c}{196.60 $\pm$ 10.13} & 16.25 & \multicolumn{2}{c}{-2653.42 $\pm$ 0.63} &3.02 & \multicolumn{2}{c}{-419973 $\pm$ 1.11} & 1.41 &\multicolumn{2}{c}{-3522.67 $\pm$ 0.80} & 1.81\\ 	
			R-LNS  & \multicolumn{2}{c}{188.02 $\pm$ 7.13} & 11.24 & \multicolumn{2}{c}{-2683.30 $\pm$ 0.59} &1.93 & \multicolumn{2}{c}{-427478 $\pm$ 0.95} & 0.61 & \multicolumn{2}{c}{-3521.90 $\pm$ 0.82} & 1.83\\ 
			FT-LNS  & \multicolumn{2}{c}{179.40 $\pm$ 8.47} & 6.04 & \multicolumn{2}{c}{-2684.94 $\pm$ 0.81} &1.87 & \multicolumn{2}{c}{-424052 $\pm$ 0.96} & 1.41 &
			\multicolumn{2}{c}{-3525.28 $\pm$ 0.83} & 1.74\\ 	
			Ours  & \multicolumn{2}{c}{176.84 $\pm$ 7.42} & 4.57 & \multicolumn{2}{c}{\textbf{-2735.86 $\pm$ 0.50}} & \textbf{0.00} & \multicolumn{2}{c}{\textbf{-428052 $\pm$ 0.12}} & \textbf{0.49} &\multicolumn{2}{c}{\textbf{-3587.72 $\pm$ 0.76}} & \textbf{0.00}\\
		\bottomrule          
	\end{tabular}}
	\end{threeparttable}
	\label{tab:generalization}
\end{table}

Training deep models that perform well on larger problems is a desirable property for solving IPs, since practical problems are often large-scale. 
Here we evaluate such generalization performance on instances with the average sizes listed in Table \ref{tab:pro_size}. We run our method and baselines on these instances with the same time limits as those in the experiments for comparative analysis. For our method and FT-LNS, we directly apply the policies trained in Section \ref{sec:exp_one}. 

All results are displayed in Table \ref{tab:generalization}. It is revealed that with the same 200s time limit, while the LNS baselines only outperform SCIP on specific problems, our LNS policies trained on small instances are consistently superior to all baselines, showing a stronger generalization ability. Also, our method delivers much smaller gaps, e.g., at least $38.35\%$ smaller than that of SCIP (200s), which are more prominent than the results in Section~\ref{sec:exp_one}. It indicates that our policies are more efficient in improving SCIP for larger instances solely by generalization. When SCIP runs with 500s, it surpasses all three LNS baselines on SC$_{2}$, CA$_{2}$ and SC$_{4}$, while our method can still deliver better results on all problems. Compared to SCIP with 1000s, our method is inferior on SC$_{2}$, CA$_{2}$ and SC$_{4}$ but apparently better on the remaining 5 instance groups. 

We further test all methods with 500s time limit except SCIP, which is allowed to run with 1000s. All results are gathered in Table \ref{tab:gene500}. It is revealed that our method still has clear advantages over others on all problems, and consistently outperforms SCIP with mere 1/2 runtime. We find that LNS baselines can outperform SCIP on some problems, especially the three largest ones, i.e., MIS$_4$, CA$_4$ and MC$_4$. It suggests that as the problem size becomes larger, LNS could be more effective to deliver high-quality solutions by solving successive sub-IPs which have much lower complexity than the original problem. On the other hand, our method outperforms all LNS baselines, showing that it is more efficient in improving the solution. In summary, our LNS policies learned on small instances generalize well to larger ones, with a persistent advantage over other methods.

\begin{table}[!t]  \small
\setlength{\tabcolsep}{5pt}
	\centering
	\caption{Generalization to large instances ($500$s).}
	\begin{threeparttable}
\scalebox{0.89}{    
	\begin{tabular}{ccccccccccccc}
		\toprule
		& \multicolumn{3}{c}{SC$_{2}$} & \multicolumn{3}{c}{MIS$_{2}$}   & \multicolumn{3}{c}{CA$_{2}$} & \multicolumn{3}{c}{MC$_{2}$}  \\
		\cmidrule(lr){2-4}
		\cmidrule(lr){5-7}\cmidrule(lr){8-10}\cmidrule(lr){11-13}
			Methods & \multicolumn{2}{c}{Obj.$\pm$Std.\%} & Gap\% & \multicolumn{2}{c}{Obj.$\pm$Std.\%} &Gap\% & \multicolumn{2}{c}{Obj.$\pm$Std.\%} & Gap\% & \multicolumn{2}{c}{Obj.$\pm$Std.\%} & Gap\%  \\ \midrule
			$\text{SCIP}^{**}$  & \multicolumn{2}{c}{295.70 $\pm$ 7.89} & 4.48 & \multicolumn{2}{c}{-1361.98 $\pm$ 1.06} & 0.56 & \multicolumn{2}{c}{-217271 $\pm$ 1.93} & 1.59 & \multicolumn{2}{c}{-1714.71 $\pm$ 1.02} & 5.44 \\ 
			U-LNS & \multicolumn{2}{c}{302.94 $\pm$ 8.15} & 7.04 & \multicolumn{2}{c}{-1368.58 $\pm$ 0.72} & 0.05  & \multicolumn{2}{c}{-200256 $\pm$ 2.08} & 9.28 & \multicolumn{2}{c}{-1777.98 $\pm$ 1.01} & 1.95 \\ 	
			R-LNS  &\multicolumn{2}{c}{298.24 $\pm$ 7.43}  & 5.39 & \multicolumn{2}{c}{-1362.04 $\pm$ 0.71}  & 0.52 & \multicolumn{2}{c}{-207937 $\pm$ 1.44} & 5.81 & \multicolumn{2}{c}{-1776.44 $\pm$ 1.02} & 2.04 \\ 
			FT-LNS  & \multicolumn{2}{c}{303.34 $\pm$ 7.97} & 7.18 & \multicolumn{2}{c}{-1345.60 $\pm$ 0.83} & 1.76  & \multicolumn{2}{c}{-213464 $\pm$ 1.21} & 3.30 & \multicolumn{2}{c}{-1767.81 $\pm$ 1.04} & 2.51 \\ 	
			Ours  &\multicolumn{2}{c}{\textbf{295.36 $\pm$ 7.81}}  & \textbf{4.36} & \multicolumn{2}{c}{\textbf{-1368.68 $\pm$ 0.65}}  & \textbf{0.04} & \multicolumn{2}{c}{\textbf{-218920 $\pm$ 2.13}} & \textbf{0.85} & \multicolumn{2}{c}{\textbf{-1813.02 $\pm$ 0.91}} & \textbf{0.02}\\ \midrule
		& \multicolumn{3}{c}{SC$_{4}$} & \multicolumn{3}{c}{MIS$_{4}$}   & \multicolumn{3}{c}{CA$_{4}$} & \multicolumn{3}{c}{MC$_{4}$}   \\
		\cmidrule(lr){2-4}
		\cmidrule(lr){5-7}\cmidrule(lr){8-10}\cmidrule(lr){11-13}
			Methods & \multicolumn{2}{c}{Obj.$\pm$Std.\%} & Gap\% & \multicolumn{2}{c}{Obj.$\pm$Std.\%} &Gap\% & \multicolumn{2}{c}{Obj.$\pm$Std.\%} & Gap\% & \multicolumn{2}{c}{Obj.$\pm$Std.\%} & Gap\% \\ \midrule
			$\text{SCIP}^{**}$  & \multicolumn{2}{c}{175.38 $\pm$ 6.99} & 5.21 & \multicolumn{2}{c}{-2673.64 $\pm$ 1.54} &2.41 & \multicolumn{2}{c}{-372291 $\pm$ 1.22} & 14.98 & \multicolumn{2}{c}{-3394.42 $\pm$ 0.81} & 6.04\\ 
			U-LNS & \multicolumn{2}{c}{185.62 $\pm$ 8.19 } & 11.26 & \multicolumn{2}{c}{-2737.24 $\pm$ 0.54} &0.09 & \multicolumn{2}{c}{-426480 $\pm$ 0.93} & 2.60 & \multicolumn{2}{c}{-3556.69 $\pm$ 0.80} & 1.55 \\ 	
			R-LNS  & \multicolumn{2}{c}{172.96 $\pm$ 6.43} & 3.73 & \multicolumn{2}{c}{-2736.60 $\pm$ 0.52} &0.11 & \multicolumn{2}{c}{-431786 $\pm$ 1.03} & 1.39 & \multicolumn{2}{c}{-3554.98 $\pm$ 0.80} & 1.59\\ 
			FT-LNS  & \multicolumn{2}{c}{175.20 $\pm$ 6.59} & 5.11 & \multicolumn{2}{c}{-2685.30 $\pm$ 0.81} &1.97 & \multicolumn{2}{c}{-431234 $\pm$ 0.91} & 1.52 & \multicolumn{2}{c}{-3526.24 $\pm$ 0.79} & 6.10\\ 	
			Ours  & \multicolumn{2}{c}{\textbf{172.38 $\pm$ 7.14}} & \textbf{3.36} & \multicolumn{2}{c}{\textbf{-2738.24 $\pm$ 0.50}} & \textbf{0.04} & \multicolumn{2}{c}{\textbf{-437880 $\pm$ 0.72}} & \textbf{0.00} & \multicolumn{2}{c}{\textbf{-3612.52 $\pm$ 0.74}} & \textbf{0.00} \\ 
		\bottomrule          
	\end{tabular}}
	\end{threeparttable}
	\label{tab:gene500}
\end{table}

\subsection{Experiments with Gurobi} \label{sec:gurobi}

Our LNS framework is generally applicable to any IP solver. Here we evaluate its performance by leveraging Gurobi (v9.0.3) \cite{gurobi} as the repair operator. Gurobi is a commercial solver and offers less interfaces to the internal solving process. Thus, we condense the static features of variables to mere objective coefficients, and attain the initial solution by running the solver with $2$s time limit. During training, we set the step limit $T$=$50$ with $1$s time limit for Gurobi at each step. For FT-LNS, we use the same $1$s time limit, and tune the group number to $2$ in all problems for its best performance. The remaining settings are the same as those with SCIP. During 
testing, we let all methods run with 100s time limit, roughly twice as much as that for training. To save space, we only show results of two instance groups for each problem in Table \ref{tab:gurobi}.\footnote{Specially, we observe that MIS is fairly easy for Gurobi (76 out of 100 instances can be solved optimally with average 40s). Thus, we evaluate this problem with less time limit and show the results in Appendix A.6.} 
Despite the shorter runtime, we find that Gurobi (100s) generally attains lower objective values than SCIP (200s), showing a better performance to solve IP problems. Hence, the LNS baselines lose their advantages over the solver on several problems, e.g., SC, CA and SC$_{2}$. In contrast, our method outperforms Gurobi across all problems, showing good performance on instances of both training and generalization sizes. Moreover, our method can be well applied to much larger instances and we provide this evaluation in Appendix A.7.

\begin{table}[!t] \small
\setlength{\tabcolsep}{0.3pt}
\renewcommand{\arraystretch}{1.2}
	\centering
	\caption{Results with Gurobi. The left part shows the results of inference on the testing set in SC, CA and MC; the right part shows the results of generalization to larger instances in SC$_{2}$, CA$_{2}$ and MC$_{2}$.}
	\begin{threeparttable}
    \scalebox{0.78}{
	\begin{tabular}{ccccccccccccccccccc}
		\toprule
	    & \multicolumn{3}{c}{SC} & \multicolumn{3}{c}{CA}   & \multicolumn{3}{c}{MC} & \multicolumn{3}{c}{SC$_{2}$} & \multicolumn{3}{c}{CA$_{2}$}   & \multicolumn{3}{c}{MC$_{2}$}\\
		\cmidrule(lr){2-4}\cmidrule(lr){5-7}\cmidrule(lr){8-10}\cmidrule(lr){11-13}
		\cmidrule(lr){14-16}\cmidrule(lr){17-19}
			Methods & \multicolumn{2}{c}{Obj.$\pm$Std.\%} & Gap\% & \multicolumn{2}{c}{Obj.$\pm$Std.\%} &Gap\% & \multicolumn{2}{c}{Obj.$\pm$Std.\%} & Gap\% & \multicolumn{2}{c}{Obj.$\pm$Std.\%} & Gap\% & \multicolumn{2}{c}{Obj.$\pm$Std.\%} &Gap\% & \multicolumn{2}{c}{Obj.$\pm$Std.\%} & Gap\% \\ \midrule
			Gurobi   & \multicolumn{2}{c}{554.94 $\pm$ 8.34} & 1.15 & \multicolumn{2}{c}{-111668 $\pm$ 1.96} & 1.10 & \multicolumn{2}{c}{-863.91 $\pm$ 3.77} & 3.31 & \multicolumn{2}{c}{302.52 $\pm$ 7.73} & 2.43 &  \multicolumn{2}{c}{-214271 $\pm$ 1.52} & 3.63 &  \multicolumn{2}{c}{-1652.83 $\pm$ 3.63} & 5.81\\
			U-LNS  & \multicolumn{2}{c}{562.08 $\pm$ 8.16} & 2.46 & \multicolumn{2}{c}{-110402 $\pm$ 1.67} & 2.21 & \multicolumn{2}{c}{-862.59 $\pm$ 1.75} & 3.45 & \multicolumn{2}{c}{301.48 $\pm$ 7.62} & 2.07 &  \multicolumn{2}{c}{-218986 $\pm$ 1.42} & 1.51 & \multicolumn{2}{c}{-1733.57 $\pm$ 1.22} & 1.21\\ 
			R-LNS  & \multicolumn{2}{c}{563.98 $\pm$ 8.29} & 2.81 & \multicolumn{2}{c}{-110230 $\pm$ 1.56} & 2.36 & \multicolumn{2}{c}{-860.22 $\pm$ 1.96} & 3.72 &\multicolumn{2}{c}{302.86 $\pm$ 7.41}  & 2.56 &  \multicolumn{2}{c}{-219462 $\pm$ 1.16} & 1.29 & \multicolumn{2}{c}{-1723.75 $\pm$ 1.32} & 1.77\\ 
			FT-LNS  & \multicolumn{2}{c}{564.14 $\pm$ 8.37} & 2.84 & \multicolumn{2}{c}{-110041 $\pm$ 1.56} & 2.53 & \multicolumn{2}{c}{-866.22 $\pm$ 1.65} & 3.03 & \multicolumn{2}{c}{348.50 $\pm$ 9.05} & 17.99 &  \multicolumn{2}{c}{-206189 $\pm$ 1.39} & 7.26 & \multicolumn{2}{c}{-1726.07 $\pm$ 1.19} & 1.64\\ 	
			Ours  &\multicolumn{2}{c}{\textbf{551.88 $\pm$ 8.31}}  & \textbf{0.59} & \multicolumn{2}{c}{\textbf{-111787 $\pm$ 2.60}}  & \textbf{1.00} & \multicolumn{2}{c}{\textbf{-888.97 $\pm$ 1.55}} & \textbf{0.50} &\multicolumn{2}{c}{\textbf{297.70 $\pm$ 7.40}}  & \textbf{0.80} &  \multicolumn{2}{c}{\textbf{-222346 $\pm$ 1.35}} & \textbf{0.00} & \multicolumn{2}{c}{\textbf{-1752.98 $\pm$ 1.21}} & \textbf{0.11}\\ 	
		\bottomrule          
	\end{tabular}}
	\end{threeparttable}
	\label{tab:gurobi}
\end{table}

\subsection{Testing on MIPLIB}

The mixed integer programming library (MIPLIB) \cite{miplib} contains real-world COPs from various domains. Since the instances in MIPLIB are severely diverse in problem types, structures and sizes, it is not a very suitable testing set to directly apply learning based models and thus seldom used in the related works. We evaluate our method (with Gurobi as the repair solver) on this realistic dataset, in the style of active search on each instance  \cite{Bello2017WorkshopT,khalil2017learning}, and compare it to SCIP and Gurobi. Results show that: 1) with the same 1000s time limit, our method is superior to both solvers on 24/35 instances and comparable to them on 9/35 instances; 2) our method with 1000s time limit outperforms both solvers with 3600s time limit on 13/35 instances; 3) for an open instance, we find a better solution than the best known one. More details are provided in Appendix A.8.

\section{Conclusions and future work} \label{sec:conclusion}
We propose a deep RL method to learn LNS policy for solving IP problems in bounded time. To tackle the issue of large action space, we apply action factorization to represent all potential variable subsets. On top of it, we design a parameter-sharing GNN to learn policies for each variable, and train it by a customized actor-critic algorithm. Results show that our method outperforms SCIP with much less time, and significantly surpasses LNS baselines with the same time. The learned policies also generalize well to larger problems. Furthermore, the evaluation of our method with Gurobi reveals that it can effectively improve this leading commercial solver. For limitations, since we mainly aim to refine off-the-shelf solvers for general IP problems, it is not sufficient to conclude that our method can transcend specialized and highly-optimized algorithms in different domains. In a practical view, our method could be a choice when new IP problems are produced with little expertise, or extensive dependence on domain knowledge is expected to be avoided. Also, our LNS policies are more suitable to improve solvers for large-scale problems in bounded time, but cannot provide optimality guarantee. For future work, we will apply our method to other (mixed) IP problems, and extend it by combining with other learning techniques for IPs, such as learning to branch.

\section*{Acknowledgments}
This research was conducted at Singtel Cognitive and Artificial Intelligence Lab for Enterprises (SCALE@NTU), which is a collaboration between Singapore Telecommunications Limited (Singtel) and Nanyang Technological University (NTU) that is supported by A*STAR under its Industry Alignment Fund (LOA Award number: I1701E0013). Wen Song was supported by the National Natural Science Foundation of China under Grant 62102228, and the Young Scholar Future Plan of Shandong University under Grant 62420089964188. Zhiguang Cao was supported by the National Natural Science Foundation of China under Grant 61803104.


\newpage
\bibliographystyle{unsrtnat}
\bibliography{nips21}

\begin{thebibliography}{48}
\providecommand{\natexlab}[1]{#1}
\providecommand{\url}[1]{\texttt{#1}}
\expandafter\ifx\csname urlstyle\endcsname\relax
  \providecommand{\doi}[1]{doi: #1}\else
  \providecommand{\doi}{doi: \begingroup \urlstyle{rm}\Url}\fi

\bibitem[Paschos(2013)]{paschos2013applications}
Vangelis~Th Paschos.
\newblock \emph{Applications of combinatorial optimization}.
\newblock John Wiley \& Sons, 2013.

\bibitem[Papadimitriou and Steiglitz(1998)]{papadimitriou1998combinatorial}
Christos~H Papadimitriou and Kenneth Steiglitz.
\newblock \emph{Combinatorial optimization: algorithms and complexity}.
\newblock Courier Corporation, 1998.

\bibitem[Bengio et~al.(2020)Bengio, Lodi, and Prouvost]{bengio2018machine}
Yoshua Bengio, Andrea Lodi, and Antoine Prouvost.
\newblock Machine learning for combinatorial optimization: a methodological
  tour d’horizon.
\newblock \emph{European Journal of Operational Research}, 290\penalty0
  (2):\penalty0 405--421, 2020.

\bibitem[Vinyals et~al.(2015)Vinyals, Fortunato, and
  Jaitly]{vinyals2015pointer}
Oriol Vinyals, Meire Fortunato, and Navdeep Jaitly.
\newblock Pointer networks.
\newblock In \emph{Proceedings of the 29th Conference on Neural Information
  Processing Systems (NIPS)}, pages 2692--2700, 2015.

\bibitem[Bello and Pham(2017)]{Bello2017WorkshopT}
Irwan Bello and Hieu Pham.
\newblock Neural combinatorial optimization with reinforcement learning.
\newblock In \emph{the 5th International Conference on Learning Representations
  (ICLR)}, 2017.

\bibitem[Dai et~al.(2017)Dai, Khalil, Zhang, Dilkina, and
  Song]{khalil2017learning}
Hanjun Dai, Elias Khalil, Yuyu Zhang, Bistra Dilkina, and Le~Song.
\newblock Learning combinatorial optimization algorithms over graphs.
\newblock In \emph{Proceedings of the 31st Conference on Neural Information
  Processing Systems (NIPS)}, pages 6348--6358, 2017.

\bibitem[Kool et~al.(2019)Kool, van Hoof, and Welling]{kool2018attention}
Wouter Kool, Herke van Hoof, and Max Welling.
\newblock Attention, learn to solve routing problems!
\newblock In \emph{the 7th International Conference on Learning Representations
  (ICLR)}, 2019.

\bibitem[Xin et~al.(2021)Xin, Song, Cao, and Zhang]{xin2020multi}
Liang Xin, Wen Song, Zhiguang Cao, and Jie Zhang.
\newblock Multi-decoder attention model with embedding glimpse for solving
  vehicle routing problems.
\newblock In \emph{Proceedings of the 35th AAAI Conference on Artificial
  Intelligence}, page 12042–12049, 2021.

\bibitem[Li et~al.(2021{\natexlab{a}})Li, Ma, Gao, Cao, Lim, Song, and
  Zhang]{li2021deep}
Jingwen Li, Yining Ma, Ruize Gao, Zhiguang Cao, Andrew Lim, Wen Song, and Jie
  Zhang.
\newblock Deep reinforcement learning for solving the heterogeneous capacitated
  vehicle routing problem.
\newblock \emph{IEEE Transactions on Cybernetics}, 2021{\natexlab{a}}.

\bibitem[Li et~al.(2021{\natexlab{b}})Li, Xin, Cao, Lim, Song, and
  Zhang]{li2021heterogeneous}
Jingwen Li, Liang Xin, Zhiguang Cao, Andrew Lim, Wen Song, and Jie Zhang.
\newblock Heterogeneous attentions for solving pickup and delivery problem via
  deep reinforcement learning.
\newblock \emph{IEEE Transactions on Intelligent Transportation Systems},
  2021{\natexlab{b}}.

\bibitem[Zhang et~al.(2020)Zhang, Song, Cao, Zhang, Tan, and
  Xu]{zhang2020learning}
Cong Zhang, Wen Song, Zhiguang Cao, Jie Zhang, Puay~Siew Tan, and Chi Xu.
\newblock Learning to dispatch for job shop scheduling via deep reinforcement
  learning.
\newblock In \emph{Proceedings of the 34th Advances in Neural Information
  Processing Systems (NIPS)}, pages 1621--1632, 2020.

\bibitem[Chen and Tian(2019)]{chen2019learning}
Xinyun Chen and Yuandong Tian.
\newblock Learning to perform local rewriting for combinatorial optimization.
\newblock In \emph{Proceedings of the 33rd Conference on Neural Information
  Processing Systems (NIPS)}, pages 6278--6289, 2019.

\bibitem[Wu et~al.(2021)Wu, Song, Cao, Zhang, and Lim]{wu2019learning}
Yaoxin Wu, Wen Song, Zhiguang Cao, Jie Zhang, and Andrew Lim.
\newblock Learning improvement heuristics for solving routing problems.
\newblock \emph{IEEE Transactions on Neural Networks and Learning Systems},
  2021.

\bibitem[Lu et~al.(2020)Lu, Zhang, and Yang]{lu2019learning}
Hao Lu, Xingwen Zhang, and Shuang Yang.
\newblock A learning-based iterative method for solving vehicle routing
  problems.
\newblock In \emph{the 8th International Conference on Learning Representations
  (ICLR)}, 2020.

\bibitem[Hottung and Tierney(2020)]{hottung2019neural}
Andr{\'e} Hottung and Kevin Tierney.
\newblock Neural large neighborhood search for the capacitated vehicle routing
  problem.
\newblock In \emph{Proceedings of the 24th European Conference on Artificial
  Intelligence (ECAI)}, 2020.

\bibitem[Gao et~al.(2020)Gao, Chen, Chen, Luo, Zhu, and Liu]{gao2020learn}
Lei Gao, Mingxiang Chen, Qichang Chen, Ganzhong Luo, Nuoyi Zhu, and Zhixin Liu.
\newblock Learn to design the heuristics for vehicle routing problem.
\newblock \emph{arXiv preprint arXiv:2002.08539}, 2020.

\bibitem[Land and Doig(2010)]{land2010automatic}
Ailsa~H Land and Alison~G Doig.
\newblock An automatic method for solving discrete programming problems.
\newblock In \emph{50 Years of Integer Programming 1958-2008}, pages 105--132.
  Springer, 2010.

\bibitem[He et~al.(2014)He, Daume~III, and Eisner]{he2014learning}
He~He, Hal Daume~III, and Jason~M Eisner.
\newblock Learning to search in branch and bound algorithms.
\newblock In \emph{Proceedings of the 27th Conference on Neural Information
  Processing Systems (NIPS)}, pages 3293--3301, 2014.

\bibitem[Khalil et~al.(2017)Khalil, Dilkina, Nemhauser, Ahmed, and
  Shao]{khalil2017learn}
Elias~B Khalil, Bistra Dilkina, George~L Nemhauser, Shabbir Ahmed, and Yufen
  Shao.
\newblock Learning to run heuristics in tree search.
\newblock In \emph{Proceedings of the 26th International Joint Conference on
  Artificial Intelligence (IJCAI)}, pages 659--666, 2017.

\bibitem[Gasse et~al.(2019)Gasse, Ch{\'e}telat, Ferroni, Charlin, and
  Lodi]{gasse2019exact}
Maxime Gasse, Didier Ch{\'e}telat, Nicola Ferroni, Laurent Charlin, and Andrea
  Lodi.
\newblock Exact combinatorial optimization with graph convolutional neural
  networks.
\newblock In \emph{Proceedings of the 33rd Conference on Neural Information
  Processing Systems (NIPS)}, 2019.

\bibitem[Song et~al.(2020)Song, Lanka, Yue, and Dilkina]{song2020general}
Jialin Song, Ravi Lanka, Yisong Yue, and Bistra Dilkina.
\newblock A general large neighborhood search framework for solving integer
  linear programs.
\newblock In \emph{Proceedings of the 34th Conference on Neural Information
  Processing Systems (NIPS)}, 2020.

\bibitem[Tang et~al.(2020)Tang, Agrawal, and Faenza]{tang2020reinforcement}
Yunhao Tang, Shipra Agrawal, and Yuri Faenza.
\newblock Reinforcement learning for integer programming: Learning to cut.
\newblock In \emph{International Conference on Machine Learning (ICML)}, pages
  9367--9376, 2020.

\bibitem[Ding et~al.(2020)Ding, Zhang, Shen, Li, Wang, Xu, and
  Song]{ding2020accelerating}
Jian-Ya Ding, Chao Zhang, Lei Shen, Shengyin Li, Bing Wang, Yinghui Xu, and
  Le~Song.
\newblock Accelerating primal solution findings for mixed integer programs
  based on solution prediction.
\newblock In \emph{Proceedings of the AAAI Conference on Artificial
  Intelligence}, volume~34, pages 1452--1459, 2020.

\bibitem[Nair et~al.(2020)Nair, Bartunov, Gimeno, von Glehn, Lichocki, Lobov,
  O'Donoghue, Sonnerat, Tjandraatmadja, Wang, et~al.]{nair2020solving}
Vinod Nair, Sergey Bartunov, Felix Gimeno, Ingrid von Glehn, Pawel Lichocki,
  Ivan Lobov, Brendan O'Donoghue, Nicolas Sonnerat, Christian Tjandraatmadja,
  Pengming Wang, et~al.
\newblock Solving mixed integer programs using neural networks.
\newblock \emph{arXiv preprint arXiv:2012.13349}, 2020.

\bibitem[Pazis and Parr(2011)]{pazis2011generalized}
Jason Pazis and Ronald Parr.
\newblock Generalized value functions for large action sets.
\newblock In \emph{Proceedings of the 28th International Conference on Machine
  Learning (ICML)}, pages 1185--1192, 2011.

\bibitem[Tavakoli et~al.(2018)Tavakoli, Pardo, and
  Kormushev]{tavakoli2018action}
Arash Tavakoli, Fabio Pardo, and Petar Kormushev.
\newblock Action branching architectures for deep reinforcement learning.
\newblock In \emph{Proceedings of the AAAI Conference on Artificial
  Intelligence}, volume~32, 2018.

\bibitem[Dulac-Arnold et~al.(2015)Dulac-Arnold, Evans, van Hasselt, Sunehag,
  and Lillicrap]{dulac2015deep}
Gabriel Dulac-Arnold, Richard Evans, Hado van Hasselt, Peter Sunehag, and
  Timothy Lillicrap.
\newblock Deep reinforcement learning in large discrete action spaces.
\newblock \emph{arXiv preprint arXiv:1512.07679}, 2015.

\bibitem[Chandak et~al.(2019)Chandak, Theocharous, Kostas, Jordan, and
  Thomas]{chandak2019learning}
Yash Chandak, Georgios Theocharous, James Kostas, Scott Jordan, and Philip
  Thomas.
\newblock Learning action representations for reinforcement learning.
\newblock In \emph{Proceedings of the 36th International Conference on Machine
  Learning (ICML)}, pages 941--950, 2019.

\bibitem[Tang and Agrawal(2020)]{tang2020discretizing}
Yunhao Tang and Shipra Agrawal.
\newblock Discretizing continuous action space for on-policy optimization.
\newblock In \emph{Proceedings of the AAAI Conference on Artificial
  Intelligence}, volume~34, pages 5981--5988, 2020.

\bibitem[Hubert et~al.(2021)Hubert, Schrittwieser, Antonoglou, Barekatain,
  Schmitt, and Silver]{hubert2021learning}
Thomas Hubert, Julian Schrittwieser, Ioannis Antonoglou, Mohammadamin
  Barekatain, Simon Schmitt, and David Silver.
\newblock Learning and planning in complex action spaces.
\newblock \emph{arXiv preprint arXiv:2104.06303}, 2021.

\bibitem[Andrychowicz et~al.(2020)Andrychowicz, Baker, Chociej, Jozefowicz,
  McGrew, Pachocki, Petron, Plappert, Powell, Ray,
  et~al.]{andrychowicz2020learning}
OpenAI:~Marcin Andrychowicz, Bowen Baker, Maciek Chociej, Rafal Jozefowicz, Bob
  McGrew, Jakub Pachocki, Arthur Petron, Matthias Plappert, Glenn Powell, Alex
  Ray, et~al.
\newblock Learning dexterous in-hand manipulation.
\newblock \emph{The International Journal of Robotics Research}, 39\penalty0
  (1):\penalty0 3--20, 2020.

\bibitem[Pisinger and Ropke(2010)]{pisinger2010large}
David Pisinger and Stefan Ropke.
\newblock Large neighborhood search.
\newblock In \emph{Handbook of metaheuristics}, pages 399--419. Springer, 2010.

\bibitem[Ropke and Pisinger(2006)]{ropke2006adaptive}
Stefan Ropke and David Pisinger.
\newblock An adaptive large neighborhood search heuristic for the pickup and
  delivery problem with time windows.
\newblock \emph{Transportation science}, 40\penalty0 (4):\penalty0 455--472,
  2006.

\bibitem[Prescott-Gagnon et~al.(2009)Prescott-Gagnon, Desaulniers, and
  Rousseau]{prescott2009branch}
Eric Prescott-Gagnon, Guy Desaulniers, and Louis-Martin Rousseau.
\newblock A branch-and-price-based large neighborhood search algorithm for the
  vehicle routing problem with time windows.
\newblock \emph{Networks: An International Journal}, 54\penalty0 (4):\penalty0
  190--204, 2009.

\bibitem[Matignon et~al.(2012)Matignon, Laurent, and
  Le~Fort-Piat]{matignon2012independent}
Laetitia Matignon, Guillaume~J Laurent, and Nadine Le~Fort-Piat.
\newblock Independent reinforcement learners in cooperative markov games: a
  survey regarding coordination problems.
\newblock \emph{The Knowledge Engineering Review}, 27\penalty0 (1):\penalty0
  1--31, 2012.

\bibitem[Battaglia et~al.(2018)Battaglia, Hamrick, Bapst, Sanchez-Gonzalez,
  Zambaldi, Malinowski, Tacchetti, Raposo, Santoro, Faulkner,
  et~al.]{battaglia2018relational}
Peter~W Battaglia, Jessica~B Hamrick, Victor Bapst, Alvaro Sanchez-Gonzalez,
  Vinicius Zambaldi, Mateusz Malinowski, Andrea Tacchetti, David Raposo, Adam
  Santoro, Ryan Faulkner, et~al.
\newblock Relational inductive biases, deep learning, and graph networks.
\newblock \emph{arXiv preprint arXiv:1806.01261}, 2018.

\bibitem[Wu et~al.(2012)Wu, Pan, Chen, Long, Zhang, and
  Philip]{wu2020comprehensive}
Zonghan Wu, Shirui Pan, Fengwen Chen, Guodong Long, Chengqi Zhang, and S~Yu
  Philip.
\newblock A comprehensive survey on graph neural networks.
\newblock \emph{IEEE transactions on neural networks and learning systems},
  32\penalty0 (1):\penalty0 4--24, 2012.

\bibitem[Kipf and Welling(2017)]{kipf2017semi}
Thomas~N. Kipf and Max Welling.
\newblock Semi-supervised classification with graph convolutional networks.
\newblock In \emph{the 5th International Conference on Learning Representations
  (ICLR)}, 2017.

\bibitem[Duvenaud et~al.(2015)Duvenaud, Maclaurin, Aguilera-Iparraguirre,
  G{\'o}mez-Bombarelli, Hirzel, Aspuru-Guzik, and
  Adams]{duvenaud2015convolutional}
David Duvenaud, Dougal Maclaurin, Jorge Aguilera-Iparraguirre, Rafael
  G{\'o}mez-Bombarelli, Timothy Hirzel, Al{\'a}n Aspuru-Guzik, and Ryan~P
  Adams.
\newblock Convolutional networks on graphs for learning molecular fingerprints.
\newblock In \emph{Proceedings of the 29th Conference on Neural Information
  Processing Systems (NIPS)}, pages 2224--2232, 2015.

\bibitem[Yao et~al.(2019)Yao, Mao, and Luo]{yao2019graph}
Liang Yao, Chengsheng Mao, and Yuan Luo.
\newblock Graph convolutional networks for text classification.
\newblock In \emph{Proceedings of the AAAI conference on artificial
  intelligence}, volume~33, pages 7370--7377, 2019.

\bibitem[Williams(1992)]{williams1992simple}
Ronald~J Williams.
\newblock Simple statistical gradient-following algorithms for connectionist
  reinforcement learning.
\newblock \emph{Machine learning}, 8\penalty0 (3-4):\penalty0 229--256, 1992.

\bibitem[Balas and Ho(1980)]{balas1980set}
Egon Balas and Andrew Ho.
\newblock Set covering algorithms using cutting planes, heuristics, and
  subgradient optimization: a computational study.
\newblock In \emph{Combinatorial Optimization}, pages 37--60. Springer, 1980.

\bibitem[Bergman et~al.(2016)Bergman, Cire, Van~Hoeve, and
  Hooker]{bergman2016decision}
David Bergman, Andre~A Cire, Willem-Jan Van~Hoeve, and John Hooker.
\newblock \emph{Decision diagrams for optimization}.
\newblock Springer, 2016.

\bibitem[Leyton-Brown et~al.(2000)Leyton-Brown, Pearson, and
  Shoham]{leyton2000towards}
Kevin Leyton-Brown, Mark Pearson, and Yoav Shoham.
\newblock Towards a universal test suite for combinatorial auction algorithms.
\newblock In \emph{Proceedings of the 2nd ACM conference on Electronic
  commerce}, pages 66--76, 2000.

\bibitem[Albert and Barab{\'a}si(2002)]{albert2002statistical}
R{\'e}ka Albert and Albert-L{\'a}szl{\'o} Barab{\'a}si.
\newblock Statistical mechanics of complex networks.
\newblock \emph{Reviews of modern physics}, 74\penalty0 (1):\penalty0 47, 2002.

\bibitem[Achterberg(2009)]{achterberg2009scip}
Tobias Achterberg.
\newblock Scip: solving constraint integer programs.
\newblock \emph{Mathematical Programming Computation}, 1\penalty0 (1):\penalty0
  1--41, 2009.

\bibitem[Gurobi~Optimization(2021)]{gurobi}
LLC Gurobi~Optimization.
\newblock Gurobi optimizer reference manual, 2021.
\newblock URL \url{http://www.gurobi.com}.

\bibitem[Gleixner et~al.(2021)Gleixner, Hendel, Gamrath, Achterberg, Bastubbe,
  Berthold, Christophel, Jarck, Koch, Linderoth, L\"ubbecke, Mittelmann,
  Ozyurt, Ralphs, Salvagnin, and Shinano]{miplib}
Ambros Gleixner, Gregor Hendel, Gerald Gamrath, Tobias Achterberg, Michael
  Bastubbe, Timo Berthold, Philipp~M. Christophel, Kati Jarck, Thorsten Koch,
  Jeff Linderoth, Marco L\"ubbecke, Hans~D. Mittelmann, Derya Ozyurt, Ted~K.
  Ralphs, Domenico Salvagnin, and Yuji Shinano.
\newblock {MIPLIB 2017: Data-Driven Compilation of the 6th Mixed-Integer
  Programming Library}.
\newblock \emph{Mathematical Programming Computation}, 2021.
\newblock \doi{10.1007/s12532-020-00194-3}.
\newblock URL \url{https://doi.org/10.1007/s12532-020-00194-3}.

\end{thebibliography}

\newpage
\appendix

\setcounter{figure}{0} \renewcommand{\thefigure}{A.\arabic{figure}}
\setcounter{table}{0} \renewcommand{\thetable}{A.\arabic{table}}
\setcounter{section}{0} \renewcommand{\thesection}{A.\arabic{section}}


\vbox{
\hrule height 4pt
\vskip 0.25in
\vskip -\parskip%
\centering
{\LARGE\bf Learning Large Neighborhood Search Policy for Integer Programming (Appendix) }
\vskip 0.29in
\vskip -\parskip
\hrule height 1pt
}

\section{Architecture of bipartite GCN}

\begin{figure}[hbt!]
   \centering
   \includegraphics[width=1\textwidth, trim = 1.6cm 0cm 1.3cm 0cm, clip]{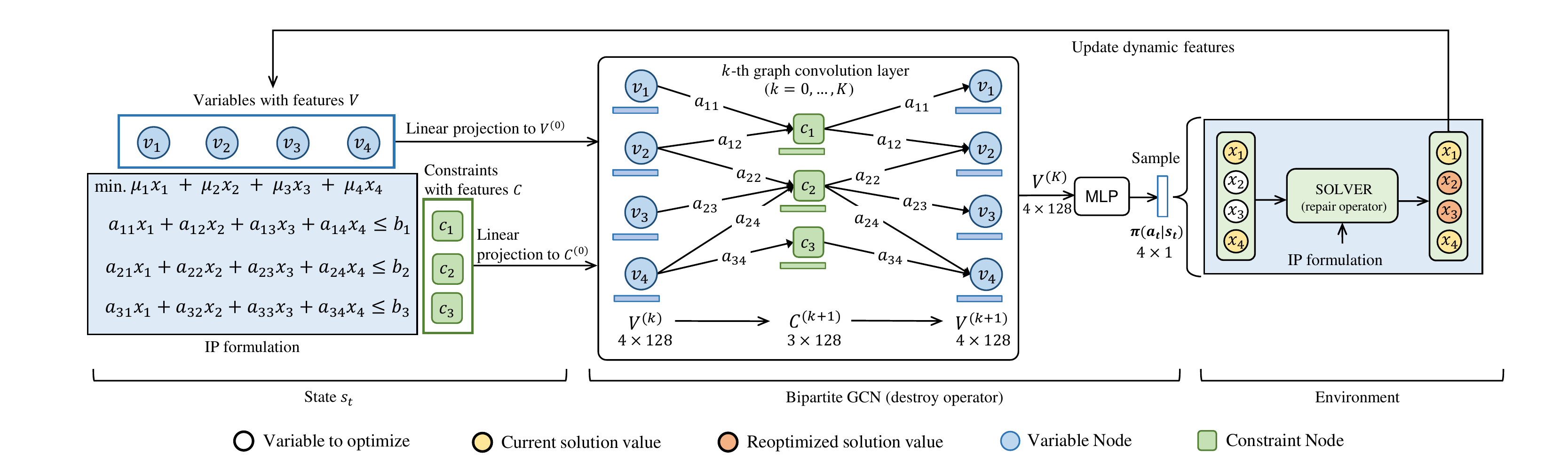}
    \vspace{-0.45cm}
   \caption{Illustration of our LNS framework with the bipartite GCN based destroy operator.}
    \label{fig:net}
\end{figure}

In this paper, we propose to factorize the selection of a variable subset into decisions on selection of each variable, under our LNS framework. To represent such action factorization, we employ the bipartite GCN as the destroy operator, as shown in Figure~\ref{fig:net}. In specific, the bipartite GCN comprises two stacks of graph convolution layers to compute the embeddings of variables, and one MLP module that computes probabilities of selecting each variable in parallel. 

\section{Training details}

Our RL algorithm for training LNS policies is depicted by the pseudo code in Algorithm \ref{alg:training}. Compared to the standard actor-critic algorithm, we use experience replay to empower the reuse of past samples (lines 2-8). In addition, we customize the standard Q-actor-critic algorithm for the proposed action factorization, by specializing the loss functions.
\IncMargin{0.5em}
\begin{algorithm}[htb!] \small
    \caption{Customized Q-actor-critic for LNS} \label{alg:training}
    \KwIn{actor $\pi_\theta$ with parameters $\theta$; critic $Q_\omega$ with parameters $\omega$; empty reply buffer $\mathcal{D}$; number of iterations $\mathit{J}$; step limit $\mathit{T}$; number of updates $U$; learning rates $\alpha_\theta$, $\alpha_\omega$; discount factor $\gamma$.}
    \For{$\mathit{j}$ = 1, 2, $\cdots$, $\mathit{J}$}{
          draw $\mathit{M}$ training instances\;
          \For{$\mathit{m}$ = 1, 2, $\cdots$, $\mathit{M}$}{
               \For{$\mathit{t}$ = 1, 2, $\cdots$, $\mathit{T}$}{
                    sample $a^i_t\sim\pi_{\theta}(a^i_t|s_t)$, derive the union $a_t$ \;
                    receive reward $r_t$ and next state $s_{t+1}$\;
                    sample $a^i_{t+1}\sim\pi_{\theta}(a^i_{t+1}|s_{t+1})$, derive $a_{t+1}$\; 
                    store transition $(s_t,a_t,r_t,s_{t+1},a_{t+1})$ in $\mathcal{D}$\; 
             }
         }
          \For{$\mathit{u}$ = 1, 2, $\cdots$, $\mathit{U}$}{
               randomly sample a batch of transitions $\mathcal{B}$ from $\mathcal{D}$\;
               update the parameters of actor and critic with
               $y_t = \gamma Q_{\omega}(s_{t+1},a_{t+1})+r_t$;
               $\omega\leftarrow\omega + \alpha_\omega\nabla_\omega\frac{1}{|\mathcal{B}|}\sum_{\mathcal{B}} (y_t-Q_\omega(s_t,a_t))^2$; 
               $z_t=\sum_{i=1}^{n}\log(\pi_{\theta}(a_t^i|s_t))$;
               $\theta\leftarrow\theta + \alpha_\theta\nabla_\theta\frac{1}{|\mathcal{B}|}\sum_{\mathcal{B}} Q_\omega(s_t,a_t)z_t$\;
         }         
    }
\end{algorithm}
\DecMargin{0.5em}

\section{Semi-shared vs. fully-shared network}

\begin{wrapfigure}{r}{0.5\textwidth}
      \centering
        \includegraphics[width=.5\textwidth]{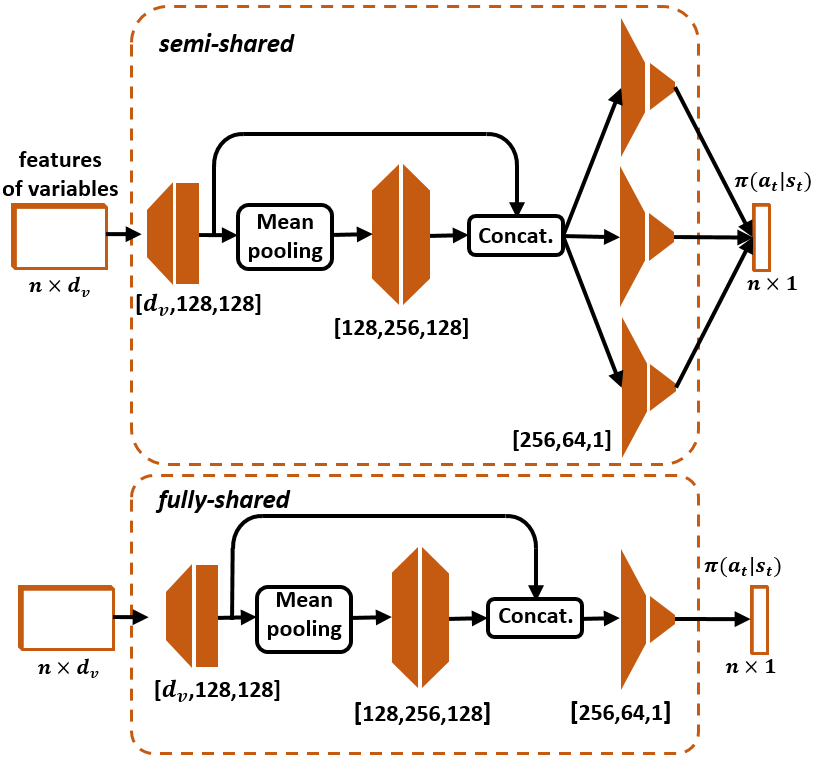}
      \caption{Architectures of the semi-shared and fully-shared policy networks.}
      \label{fig:appnetwork}
\end{wrapfigure}

As stated in the main paper, our method can be employed to train any kind of policy network that is able to represent the factorized action. We choose the fully-shared neural network since it can be generalized to different problem sizes, which is critical to solve IP problems. Nevertheless, we also experiment with a semi-shared neural network to show its performance on problem instances of the fixed size. The architecture of the neural network is displayed in the upper half of Figure~\ref{fig:appnetwork}, which is similar to the one used for DQN based RL algorithms in \cite{tavakoli2018action}. Specifically, given a collection of features for each variable, we first process them by a MLP to obtain variable embeddings. Then, these embeddings are averaged and advanced by another MLP to obtain the state embedding. Lastly, we concatenate each variable embedding with the state embedding and pass them through $n$ MLPs separately to get probabilities of $n$ variables. For the fully-shared counterpart, we only replace the $n$ MLPs by a parameter-sharing MLP, as shown in the lower half of Figure~\ref{fig:appnetwork}. All layers in MLPs are activated by Tanh except the final output by Sigmoid.

\begin{figure*}[hbt!]
\centering
\begin{tabular}{@{}c@{}c@{}c@{}}
  \subfigure[SC]
  {\includegraphics[width=0.27\linewidth]{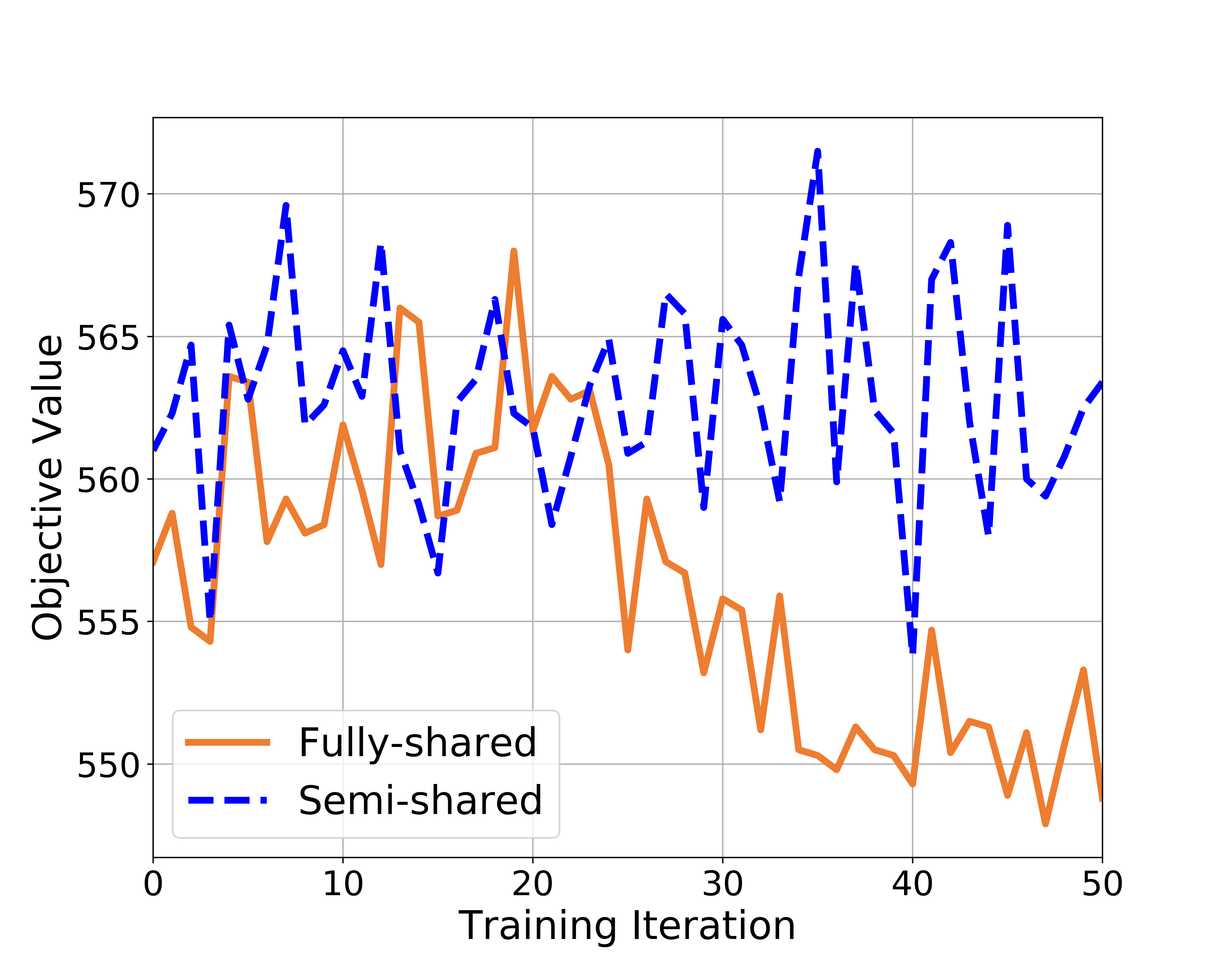}}\hspace{-3.5mm}
  \subfigure[MIS]
  {\includegraphics[width=0.27\linewidth]{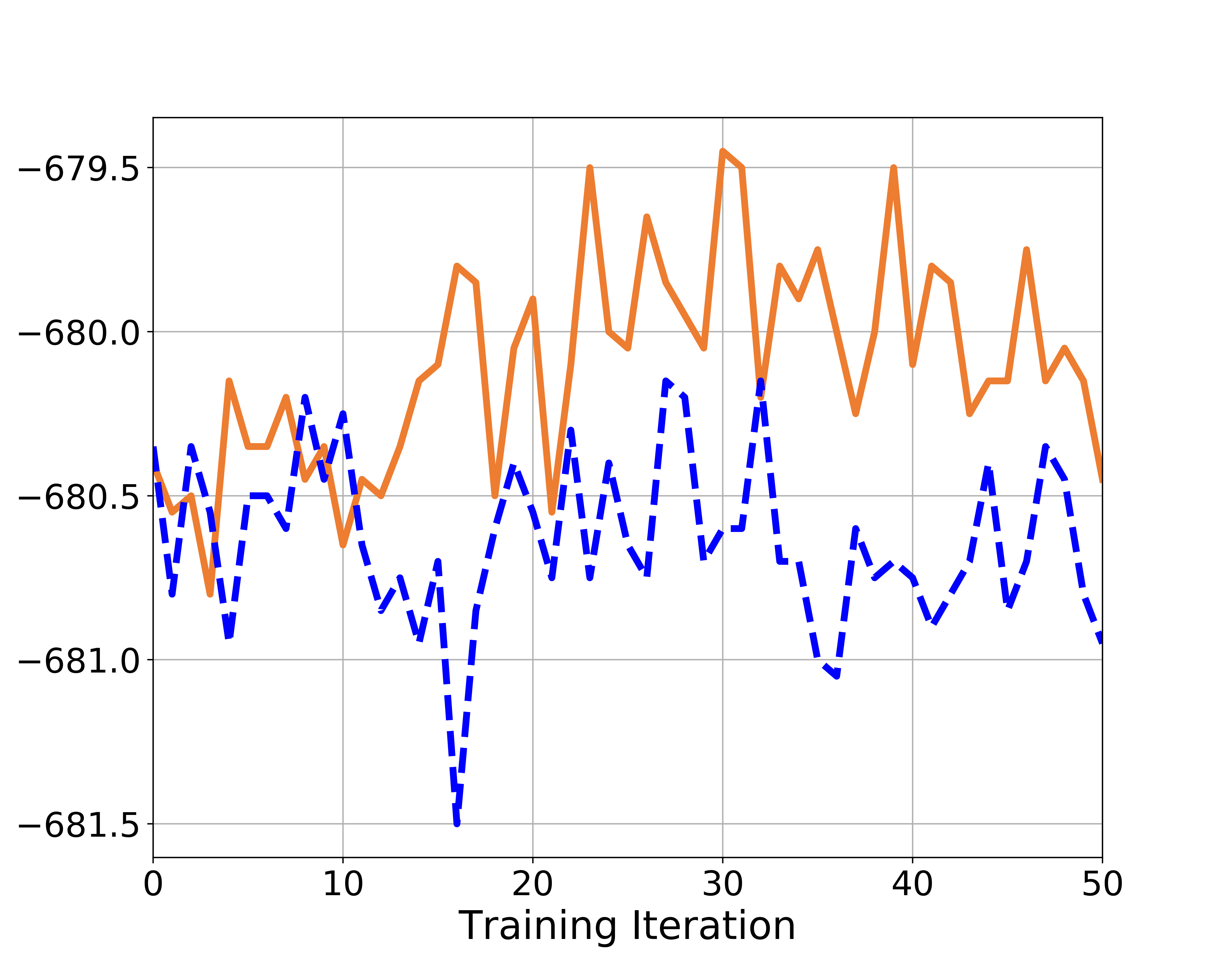}}\hspace{-3.5mm}
  \subfigure[CA]
  {\includegraphics[width=0.27\linewidth]{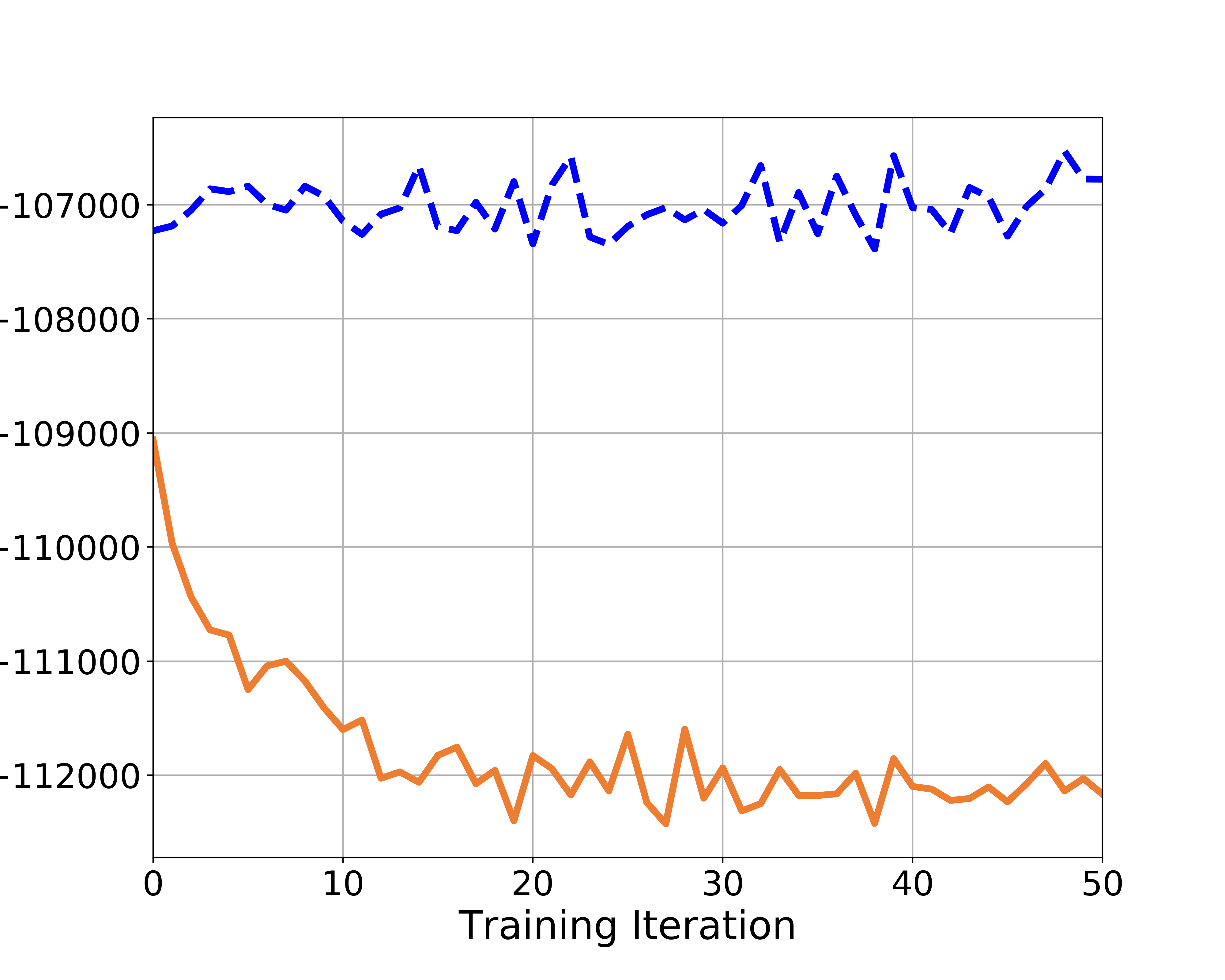}}\hspace{-3.5mm}
  \subfigure[MC]
  {\includegraphics[width=0.27\linewidth]{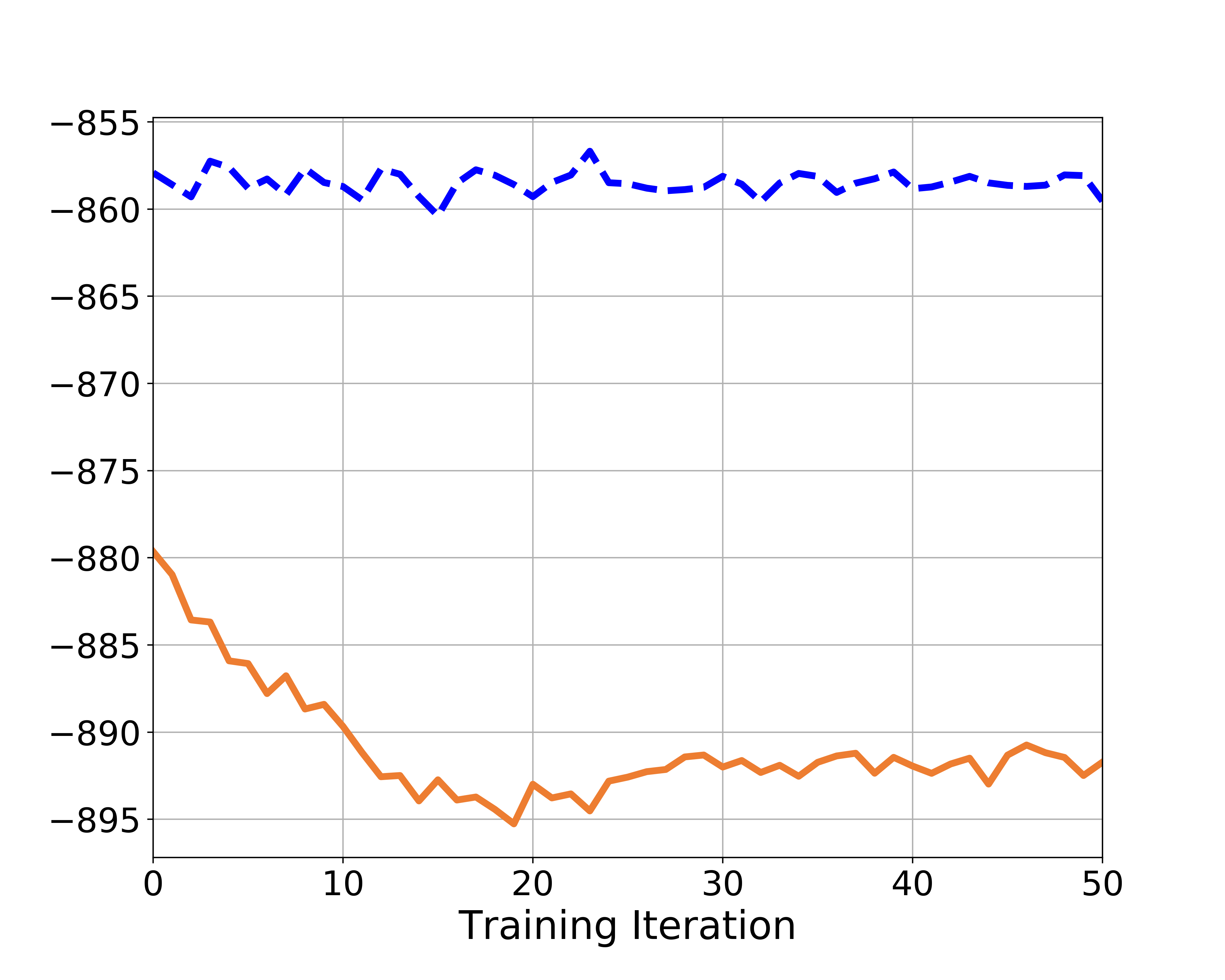}}
\end{tabular}
\caption{Training curves of the semi-shared and fully-shared networks.}
\label{fig:curve} 
\end{figure*}

Following the experimental setting for SCIP in the main paper, we train both the fully-shared and semi-shared networks on SC, MIS, CA and MC, with the customized actor-critic algorithm we designed. We evaluate the average objective value over the validation set after each training iteration. All training curves of initial $50$ iterations are displayed in Figure~\ref{fig:curve}. We find that the fully-shared network is able to learn efficiently on SC, CA and MC, while semi-shared network performs better on MIS. It indicates that our training algorithm can be applied to different kinds of policy networks, and the fully-shared network is more effective in learning LNS policies for IP problems. 
For the semi-shared network, despite the good performance with relatively low-dimensional action spaces in \cite{tavakoli2018action}, it needs far more sub-networks in our RL tasks with thousands of action dimensions, which are intractable to train together and also prevent the generalization to different problem sizes.

\section{State features}

In this paper, we represent the state by a bipartite graph $\mathcal{G}$ = $(\mathcal{V},\mathcal{C},\mathbf{A})$ attached by the features of variables, constraints and edges (i.e. $\mathbf{V}$, $\mathbf{C}$ and A), which are listed in Table \ref{tab:fea}. The logic behind this design is to reflect the instance information and the dynamic solving status, both of which are critical to learn effective policies in LNS. For the static features, we consider the ones used in \cite{gasse2019exact}, which learns variable selection policies in B\&B algorithm. It has been shown that these features have the potential to predict variables for branching. From the perspective of learning LNS, we only extract these features at the root node as the instance information. Also, we preprocess these features in two ways: 1) we delete the ones with zero variance from the features of variables, which are the same constant across all training instances; 2) we only use the right-hand-side (RHS) vector as the features for constraints. For dynamic features, we consider the efficiency of the solving process and directly record values of the current solution and incumbent at each step of LNS. These dynamic features are linked with the static features of variables and then attached to variable nodes $\mathcal{V}$. 

\begin{table}[htb!]\small
\renewcommand{\arraystretch}{1.1}
\centering
\caption{The list of features for variables, constraints and edges. \textbf{\textit{S.}} and \textbf{\textit{D.}} denote the static and dynamic attribute, respectively.}
  \begin{tabular}{m{3.3cm}<{\raggedright}m{7.3cm}<{\raggedright}m{1cm}<{\centering}m{0.5cm}<{\centering}}
    \toprule
    \textbf{Feature Types} & \textbf{Description} & \textbf{Length} & \textbf{\textit{S.}/\textit{D.}}\\
    \hline 
    Variable features ($\mathbf{V}$) & Normalized reduced cost. &  1 & \textbf{\textit{S.}}\\
    ~ & Normalized objective coefficient. & 1 & \textbf{\textit{S.}}\\
    ~ & Normalized LP age. & 1 & \textbf{\textit{S.}}\\
    ~ & Equality of solution value and lower bound, $0$ or $1$. & 1 & \textbf{\textit{S.}}\\
    ~ &  Equality of solution value and upper bound, $0$ or $1$. & 1 & \textbf{\textit{S.}}\\   
    ~ &  Fractionality of solution value. & 1 & \textbf{\textit{S.}}\\  
    ~ &  One-hot encoding of simplex basis status (i.e., lower, basic, upper). & 3 & \textbf{\textit{S.}}\\   
    ~ &  Solution value at root node. & 1 & \textbf{\textit{S.}}\\       
    ~ & Solution value at the current step. & 1 & \textbf{\textit{D.}}\\
    ~ & Value in the incumbent.  & 1 & \textbf{\textit{D.}}\\
    ~ & Average value in historical incumbents. & 1 & \textbf{\textit{D.}}\\
    \hline
    Constraint features ($\mathbf{C}$) & Constraint right-hand side. & 1 & \textbf{\textit{S.}}\\
    \hline
    Edge features ($A$) & Coefficient in incidence matrix. & 1 & \textbf{\textit{S.}}\\
    \bottomrule
\end{tabular}
\label{tab:fea}
\end{table}

\section{Comparison of short-term performance}

In the main paper, we found that FT-LNS cannot outperform R-LNS with the long time limit due to its poor generalization to unseen states. However, according to \cite{song2020general}, FT-LNS can outperform R-LNS with relatively short time limit, i.e. the similar runtime used in training of FT-LNS. To verify this point, we compare our method and LNS baselines with such setting. Specifically, we first test FT-LNS with the same number of LNS steps as in its training, and record its runtime. Then we test our method and other baselines using the runtime of FT-LNS as time limit. 
All results are summarized in Table \ref{tab:short}. As shown, FT-LNS can indeed surpass R-LNS on all problems, which indicates its effectiveness in short-term improvement and is consistent with \cite{song2020general}. Nevertheless, for these experiments with short time limits, it is clear that our method still consistently outperforms all LNS baselines with smaller gaps across all problems. 


\begin{table}[htb!]  
\setlength{\tabcolsep}{5pt}
	\centering
	\caption{Results with short time limits.}
	\begin{threeparttable}
    \scalebox{0.87}{
	\begin{tabular}{ccccccccccccc}
		\toprule
	    & \multicolumn{3}{c}{SC} & \multicolumn{3}{c}{MIS}   & \multicolumn{3}{c}{CA} & \multicolumn{3}{c}{MC}    \\
		\cmidrule(lr){2-4}
		\cmidrule(lr){5-7}\cmidrule(lr){8-10}\cmidrule(lr){11-13}
			Methods & \multicolumn{2}{c}{Obj.$\pm$Std.\%} & Gap\% & \multicolumn{2}{c}{Obj.$\pm$Std.\%} &Gap\% & \multicolumn{2}{c}{Obj.$\pm$Std.\%} & Gap\% & \multicolumn{2}{c}{Obj.$\pm$Std.\%} & Gap\%\\ \midrule
			SCIP   & \multicolumn{2}{c}{586.72 $\pm$9.14} & 3.38 & \multicolumn{2}{c}{-659.78$\pm$1.14} &2.23 & \multicolumn{2}{c}{-93715$\pm$2.64} & 8.34 & \multicolumn{2}{c}{-840.68$\pm$1.49} & 3.93 \\
			U-LNS  & \multicolumn{2}{c}{615.78$\pm$12.68} & 12.44 & \multicolumn{2}{c}{-660.30$\pm$1.13} & 2.15 & \multicolumn{2}{c}{-99302$\pm$2.75} & 2.77 & \multicolumn{2}{c}{-852.39$\pm$1.65} & 2.59 \\ 
			R-LNS  & \multicolumn{2}{c}{588.70$\pm$8.54} & 3.79 & \multicolumn{2}{c}{-667.54$\pm$1.14} & 1.08 & \multicolumn{2}{c}{-98705$\pm$1.93} & 3.37 & \multicolumn{2}{c}{-851.57$\pm$1.45} & 2.68 \\ 
			FT-LNS  & \multicolumn{2}{c}{578.38$\pm$9.23} & 1.89 & \multicolumn{2}{c}{-670.48$\pm$1.13} & 0.64  & \multicolumn{2}{c}{-100323$\pm$2.00} & 1.89 & \multicolumn{2}{c}{-865.01$\pm$1.72} & 1.15 \\ 	
			Ours  &\multicolumn{2}{c}{\textbf{575.80$\pm$8.94}}  & \textbf{1.46} & \multicolumn{2}{c}{\textbf{-673.78$\pm$0.11}}  & \textbf{0.15} & \multicolumn{2}{c}{\textbf{-100867$\pm$2.11}} & \textbf{1.37} & \multicolumn{2}{c}{\textbf{-872.47$\pm$1.24}} & \textbf{0.30}\\ 	
		\bottomrule          
	\end{tabular}}
	\end{threeparttable}
	\label{tab:short}
\end{table}

\section{Testing on MIS with Gurobi}
On the training set of MIS, Gurobi solves most instances optimally with 40s in average. As stated in the main paper, our method aims to improve solvers in bounded time and cannot guarantee optimality. Thus, in contrast to 100s time limit for other problems, we evaluate all methods on MIS with 20s. The other experimental settings follow those in Section~\ref{sec:gurobi}. The results are displayed in Table \ref{tab:MIS}. As shown, our method effectively improves Gurobi to achieve smaller gaps, although it is already able to deliver high-quality solutions quickly. Also, our method is consistently superior to baselines on all instances groups, demonstrating good generalization to different-sized problems.

\begin{table}[hbt!]  
\setlength{\tabcolsep}{8pt}
	\centering
	\caption{Results on MIS with Gurobi.}
	\begin{threeparttable}
    \scalebox{0.95}{	
	\begin{tabular}{cccccccccc}
		\toprule
		& \multicolumn{3}{c}{MIS}   & \multicolumn{3}{c}{MIS$_{2}$}& \multicolumn{3}{c}{MIS$_{4}$}    \\
		\cmidrule(lr){2-4}
		\cmidrule(lr){5-7}\cmidrule(lr){8-10}
			Methods & \multicolumn{2}{c}{Obj.$\pm$Std.\%} & Gap\% &  \multicolumn{2}{c}{Obj.$\pm$Std.\%} & Gap\% & \multicolumn{2}{c}{Obj.$\pm$Std.\%} & Gap\%   \\ \midrule			
			Gurobi & \multicolumn{2}{c}{682.22 $\pm$ 1.06} & 0.08 &  \multicolumn{2}{c}{-1359.54 $\pm$ 0.84} & 0.59 & \multicolumn{2}{c}{-2645.88 $\pm$ 1.10} & 2.96\\
			U-LNS & \multicolumn{2}{c}{682.02 $\pm$ 0.95} & 0.11 &  \multicolumn{2}{c}{-1359.36 $\pm$ 0.74} & 0.61 &\multicolumn{2}{c}{-2723.44 $\pm$ 0.52} & 0.12\\ 	
			R-LNS  & \multicolumn{2}{c}{681.82 $\pm$ 0.98} & 0.14 &  \multicolumn{2}{c}{-1365.64 $\pm$ 0.67} & 0.15 & \multicolumn{2}{c}{-2722.60 $\pm$ 0.52} & 0.15\\ 
			FT-LNS  & \multicolumn{2}{c}{682.20 $\pm$ 0.94} & 0.09 &  \multicolumn{2}{c}{-1358.86 $\pm$ 0.77} & 0.64 &
			\multicolumn{2}{c}{-2722.18 $\pm$ 0.56} & 0.16\\ 	
			Ours  & \multicolumn{2}{c}{\textbf{682.24 $\pm$ 0.94}} & \textbf{0.08} &  \multicolumn{2}{c}{\textbf{-1367.48 $\pm$ 0.65}} & \textbf{0.02} &\multicolumn{2}{c}{\textbf{-2724.08 $\pm$ 0.52}} & \textbf{0.09}\\
		\bottomrule          
	\end{tabular}}
	\end{threeparttable}
	\label{tab:MIS}
\end{table}

\section{Generalization with Gurobi}
Here we further evaluate the generalization of our LNS framework with Gurobi as the repair operator. We test all methods with 100s time limit, same as in Section~\ref{sec:gurobi}. For FT-LNS and our method, the learned policies on small instances are directly used. The results are summarized in Table \ref{tab:generalizationG}. We can observe that while our method is slightly inferior to U-LNS and R-LNS on SC$_{4}$, it can still generalize well to much larger instances on the problems CA$_{4}$ and MC$_{4}$, and outperform all baselines. This indicates that our method has a good generalization ability to improve Gurobi, the leading commercial solver, for solving instances of different scales.

\begin{table}[hbt!]  
\setlength{\tabcolsep}{8pt}
	\centering
	\caption{Generalization to large instances with Gurobi.}
	\begin{threeparttable}
    \scalebox{0.95}{	
	\begin{tabular}{cccccccccc}
		\toprule
		& \multicolumn{3}{c}{SC$_{4}$}   & \multicolumn{3}{c}{CA$_{4}$}& \multicolumn{3}{c}{MC$_{4}$}    \\
		\cmidrule(lr){2-4}
		\cmidrule(lr){5-7}\cmidrule(lr){8-10}
			Methods & \multicolumn{2}{c}{Obj.$\pm$Std.\%} & Gap\% &  \multicolumn{2}{c}{Obj.$\pm$Std.\%} & Gap\% & \multicolumn{2}{c}{Obj.$\pm$Std.\%} & Gap\%   \\ \midrule			
			Gurobi & \multicolumn{2}{c}{183.60 $\pm$ 7.29} & 5.30 &  \multicolumn{2}{c}{-377557 $\pm$ 0.85} & 13.87 & \multicolumn{2}{c}{-3373.11 $\pm$ 1.05} & 1.46\\
			U-LNS & \multicolumn{2}{c}{176.84 $\pm$ 6.89} & 1.43 &  \multicolumn{2}{c}{-436224 $\pm$ 1.12} & 0.49 &\multicolumn{2}{c}{-3388.31 $\pm$ 0.73} & 1.02\\ 	
			R-LNS  & \multicolumn{2}{c}{\textbf{176.08 $\pm$ 6.47}} & \textbf{1.01} &  \multicolumn{2}{c}{-435669 $\pm$ 0.67} & 0.62 & \multicolumn{2}{c}{-3389.47 $\pm$ 0.70} & 0.98\\ 
			FT-LNS  & \multicolumn{2}{c}{201.14 $\pm$ 9.65} & 15.35 &  \multicolumn{2}{c}{-395027 $\pm$ 3.56} & 9.89 &
			\multicolumn{2}{c}{-3373.20 $\pm$ 2.31} & 1.48\\ 	
			Ours  & \multicolumn{2}{c}{177.66 $\pm$ 6.65} & 1.92 &  \multicolumn{2}{c}{\textbf{-437735 $\pm$ 0.80}} & \textbf{0.15} &\multicolumn{2}{c}{\textbf{-3390.32 $\pm$ 0.73}} & \textbf{0.95}\\
		\bottomrule          
	\end{tabular}}
	\end{threeparttable}
	\label{tab:generalizationG}
\end{table}

\section{Testing on real-world instances in MIPLIB}

In this appendix, we provide details of the experiment on real-world instances in MIPLIB. These instances are grouped into "easy", "hard" and "open", according to their difficulties to solve. Since our method is more suitable for large-scale IP problems, we filter out the "easy" instances with relatively small sizes. We also filter out those instances that both SCIP and Gurobi cannot find any feasible solution with 3600s time limit, and finally choose 35 representative "hard" or "open" instances with only integer variables. Among the chosen instances, the number of variables ranges from 150 to 393800 (the average is 49563), and the number of constraints range from 301 to 850513 (the average is 96778). Also, these instances cover the typical application of COP from distinct domains, e.g., vehicle routing, cryptographic research and wireless network. To cope with such heterogeneous problems, we employ our method in the active search mode. Specifically, we apply the customized Q-actor-critic in Algorithm \ref{alg:training} to each instance, with only two instances solved in each iteration, i.e., $M=2$. In doing so, we can save computation memory and also raise the frequency of training. We use Gurobi as the repair operator and set its time limit to 2s in each LNS step. In addition, we set the step limit $T$=$100$, number of updates $U$=$10$, and batch size $\mathcal{B}$=32. For the initial solution, we use the one returned by Gurobi with 100s time limit. We set the whole time limit of active search to 1000s, and compare the results of SCIP and Gurobi with 1000s and 3600s time limits. The other settings follow those in Section~\ref{sec:gurobi}. 

All results are displayed in Table \ref{tab:miplib}. As shown, the proposed LNS framework can improve the solver effectively, and achieve better solutions than SCIP and Gurobi for most instances with the same or less runtime. Moreover, for the open instance "neos-3682128-sandon", we managed to find a new best solution.

{\scriptsize
\begin{longtable}{p{2.2cm}|ccccc|c} 
\caption{Results on MIPLIB. The "BKS" column lists the best know solutions given in MIPLIB. \textbf{Bold} and $\textbf{*}$ mean our method outperforms the solvers with 1000s and 3600s respectively. "-" means no feasible solution is found.}\label{tab:miplib}\\
\toprule
Instance & \begin{tabular}[c]{@{}c@{}}SCIP (1000s)\end{tabular} & \begin{tabular}[c]{@{}c@{}}SCIP (3600s)\end{tabular} & \begin{tabular}[c]{@{}c@{}}Gurobi (1000s)\end{tabular} & \begin{tabular}[c]{@{}c@{}}Gurobi (3600s)\end{tabular} & \begin{tabular}[c]{@{}c@{}}Ours (1000s) \end{tabular} & BKS \\
\midrule
a2864-99blp & -71 & -71 & -72 & -72 & -72 & -257  \\
bab3 & - & - & -654709.9511 & -655388.1120 & \textbf{-654912.9204}  & -656214.9542  \\
bley$\_$xs1noM & 5227928.57 & 5227928.57 & 3999391.53 & 3938322.37 & \textbf{3975481.35} & 3874310.51\\
cdc7-4-3-2 & -230 & -230 & -253 & -257 & \textbf{-276*}  & -289  \\
comp12-2idx & - & 676 & 416 & 380 & \textbf{363*} & 291 \\
ds & 509.5625 & 461.9725 & 309 & 177 & 319  & 93.52  \\
ex1010-pi & 254 & 248 & 241 & 239 & \textbf{238*}  & 235  \\
graph20-80-1rand & -1 & -1 & -3 & -6 & \textbf{-6} &  -6  \\
graph40-20-1rand & -1 & -1 & 0 & -15 & \textbf{-14} & -15  \\
neos-3426085-ticino & 234 & 232 & 226 & 226 & 226 & 225 \\
neos-3594536-henty & 410578 & 410578 & 402572 & 401948 & \textbf{402426} & 401382\\
neos-3682128-sandon & 40971070.0 & 35804070.0 & 34674767.94751 & 34666770.0 & \textbf{34666765.12338*} & 34666770\\
ns1828997 & 43 & 32 & 145 & 133 & \textbf{128*}  & 9  \\
\scalebox{.86}{nursesched-medium-hint03} & 8074 & 8074 &  144 & 115 & \textbf{131} & 115\\
opm2-z12-s8 & -36275 & -38015 & -33269 & -33269 & \textbf{-53379*}  & -58540 \\
pb-grow22 & 0.0 & 0.0 & -31152.0 & -46217.0 &  \textbf{-46881.0*}  &  -342763.0 \\
\tiny{proteindesign121hz512p9} & - & - &  1499 & 1499 & \textbf{1489*} & 1473\\
queens-30 & -33 & -39 & -39 & -39 & -39 & -40  \\
ramos3 & 242 & 242 & 252 & 245 & \textbf{248}  & 192  \\
rmine13 & -611.536750 & -611.536750 & -2854.351313 & -3493.781904 &  \textbf{-3487.807859}  & -3494.715232  \\
rmine15 & -759.289522  & -759.289522 & -192.372262 & -1979.559046 & \textbf{-5001.279118*}  & -5018.006238  \\
rococoC12-010001 & 44206.0 & 38905.0 & 35463 & 34673 & \textbf{35443} & 34270\\
s1234 & 319 & 319 & 41 & 29 & 41 & 29  \\
scpj4scip & 141 & 141 & 134 & 133 & 134 & 128 \\
scpk4 & 346 & 342 & 331 & 330 & \textbf{329*} & 321 \\
scpl4 & 296 & 296 & 281 & 279 & 281 &  262 \\
sorrell3 & -11 & -15 & -16 & -16 & -16 & -16  \\
sorrell4 & -18 & -18 & -22 & -23 & \textbf{-23} & -24 \\
sorrell7 & -152 & -152 & -183 & -187 & \textbf{-188*} & -196 \\
supportcase2 & - & - & 397 & 397 & 397 & 109137 \\
t1717 & 236546 & 228907 & 201342 & 201342 & \textbf{195894*} & 158260\\
t1722 & 138927 & 138927 & 123984 & 117171 & \textbf{117978} & 109137\\
tokyometro & - & 33134.6 &  8493.3 & 8479.5 & 8582.70 & 8329.4\\
v150d30-2hopcds & 42 & 41 & 41 & 41 & 41 & 41\\
z26 & -1029 & -1029 & -1005 & -1083 & \textbf{-1171*} & -1187\\
\bottomrule
\end{longtable}
}

\end{document}